\documentclass[lettersize,journal]{IEEEtran}
\usepackage{amsmath,amsfonts}
\usepackage{array}
\usepackage{textcomp}
\usepackage{stfloats}
\usepackage{url}
\usepackage{verbatim}

\usepackage{graphicx}
\usepackage{cite}
\usepackage{bm}
\usepackage[ruled]{algorithm2e}
\usepackage{multirow}
\usepackage{tikz}
\usepackage{url}
\newcommand*{\circled}[1]{\lower.7ex\hbox{\tikz\draw (0pt, 0pt)%
    circle (.5em) node {\makebox[1em][c]{\small #1}};}}
\usepackage{ulem}
\usepackage{enumerate}

\usepackage{subcaption}
\usepackage{lipsum}
\usepackage{pgfplots}
\usepackage{xcolor}

\usepackage[utf8]{inputenc} 
\usepackage[T1]{fontenc}    
\usepackage{booktabs}       
\usepackage{amsfonts}       
\usepackage{nicefrac}       
\usepackage{microtype}      
\usepackage[utf8]{inputenc} 
\usepackage[T1]{fontenc}    
\usepackage{graphicx}
\usepackage{balance}  
\usepackage{soul}
\usepackage{bbm}
\usepackage[utf8]{inputenc}
\usepackage{amsmath}

\usepackage{amssymb}
\usepackage{amsthm}
\usepackage{booktabs}
\usepackage{makecell}
\usetikzlibrary{matrix}
\usepackage{wrapfig}
\usepackage{hyperref}
\usepackage{enumitem}
\usepackage{float}
\usepackage{footnote}
\usepackage[textsize=small]{todonotes}

\usepackage{xspace}

\newcommand{\vldbname}{Galvatron\xspace}
\newcommand{\name}{Galvatron-BMW\xspace}

\usepackage{amsmath}

\begin{document}

\title{Improving Automatic Parallel Training via Balanced Memory Workload Optimization}



\author{
Yujie Wang,
Youhe Jiang,
Xupeng Miao, 
Fangcheng Fu, 
Shenhan Zhu,
Xiaonan Nie, 
Yaofeng Tu,
Bin Cui
\thanks{Yujie Wang, Fangcheng Fu, Shenhan Zhu, Xiaonan Nie and Youhe Jiang are with the School of CS \& Key Lab of High Confidence Software Technologies (MOE), Peking University, Beijing 100871, China. 
E-mail: \{alfredwang, ccchengff, shenhan.zhu, xiaonan.nie\}@pku.edu.cn, youhejiang@gmail.com}
\thanks{Xupeng Miao is with the Computer Science Department of Carnegie Mellon University. 
E-mail: xupeng@cmu.edu}
\thanks{Yaofeng Tu is with ZTE company.
E-mail: tu.yaofeng@zte.com.cn}
\thanks{Bin Cui is with the School of CS \& Key Lab of High Confidence Software Technologies (MOE), Peking University, Beijing 100871, and Institute of Computational Social Science, Peking University (Qingdao), China.  
E-mail: bin.cui@pku.edu.cn.}
}

\markboth{Journal of \LaTeX\ Class Files,~Vol.~14, No.~8, August~2021}%
{Shell \MakeLowercase{\textit{et al.}}: A Sample Article Using IEEEtran.cls for IEEE Journals}


\maketitle


\begin{abstract}
Transformer models have emerged as the leading approach for achieving state-of-the-art performance across various application domains, serving as the foundation for advanced large-scale deep learning (DL) models. However, efficiently training these models across multiple GPUs remains a complex challenge due to the abundance of parallelism options. Existing DL systems either require manual efforts to design distributed training plans or limit parallelism combinations to a constrained search space.
In this paper, we present \name, a novel system framework that integrates multiple prevalent parallelism dimensions and automatically identifies the most efficient hybrid parallelism strategy. To effectively navigate this vast search space, we employ a decision tree approach for decomposition and pruning based on intuitive insights. We further utilize a dynamic programming search algorithm to derive the optimal plan. Moreover, to improve resource utilization and enhance system efficiency, we propose a bi-objective optimization workflow that focuses on workload balance.
Our evaluations on different Transformer models demonstrate the capabilities of \name in automating distributed training under varying GPU memory constraints. Across all tested scenarios, \name consistently achieves superior system throughput, surpassing previous approaches that rely on limited parallelism strategies. 

\end{abstract}



\section{Introduction}
\IEEEPARstart{T}{ransformer} models have achieved great success in a wide range of deep learning (DL) applications in recent years, such as computer vision (CV)~\cite{DBLP:ViT,wei2022comparative}, natural language processing (NLP)~\cite{DBLP:conf/nips/transformer,DBLP:journals/dase/XuCDW22,Daiyi-zte,Letian-zte,DBLP:gpt3}, graph learning~\cite{graphormer,DBLP:journals/dase/LiDWHCXDL23} and recommendation systems~\cite{DBLP:BERT4REC}. For example, many Transformer variants (e.g., BERT~\cite{DBLP:conf/naacl/BERT}, GPT-2~\cite{gpt2}, T5~\cite{DBLP:googleT5}) are leading the state-of-the-art performance in various NLP tasks such as machine translation and question answering. Transformers are also applicable to image recognition (e.g, ViT~\cite{DBLP:ViT}, Swin Transformer~\cite{DBLP:conf/iccv/swin}) and multimodal tasks (e.g, CLIP~\cite{DBLP:conf/icml/clip},
DALL-E~\cite{DBLP:conf/icml/dalle}). 
Due to their superior performance, Transformers are becoming increasingly important in modern artificial intelligence industries.

Empirical evidence indicates that scaling model parameters is an effective path towards model performance improvements~\cite{DBLP:journals/corr/abs-2001-08361}. For instance, the original Transformer only has millions of model parameters while GPT-2 has 1.5 billion with superior performance~\cite{gpt2}. 
Such large amounts of parameters also incur high computational and memory costs even for emerging accelerators like GPUs.
With the increasing model scales, building and designing Transformers demand more system optimizations, and \textit{how to perform efficient Transformers training} is becoming more challenging.


Distributed DL systems adopt data and model parallelism to improve the training efficiency by utilizing multiple GPU devices. Data parallelism divides the large volume of input data into multiple parts and each device is only responsible for partial data~\cite{Zinkevich2010ParallelizedSG,DBLP:conf/nips/DeanCMCDLMRSTYN12,DBLP:conf/sigmod/MiaoNSYJM021}. It requires each device to store a whole model replica, suffering from large model scales. Model parallelism is a more promising direction that partitions the model from different \textit{parallelism dimensions} and makes each device store a subset of model parameters, such as tensor parallel~\cite{DBLP:conf/sc/megatron} and pipeline parallel~\cite{DBLP:conf/nips/gpipe,DBLP:conf/icml/NarayananPSCZ21,DBLP:conf/sosp/pipedream,DBLP:conf/mlsys/Yang00RAS21}. Various choices of the parallelism strategies lead to distinct memory consumption, communication overheads, and execution efficiency.


However, directly applying these techniques to scaling Transformers is facing crucial challenges in both system efficiency and usability. 
Some recent advanced methods have been proposed to automatically find the parallelism strategies through the fine-grained combination of data and model parallelism for individual operators in the model. For example, OptCNN~\cite{DBLP:conf/icml/optcnn}, FlexFlow~\cite{DBLP:conf/mlsys/flexflow,unger2022unity}, Tofu~\cite{DBLP:conf/eurosys/tofu}, and TensorOpt~\cite{DBLP:journals/tpds/tensoropt} consider both data and tensor parallelism and use different search algorithms to optimize the execution plans. PipeDream~\cite{DBLP:conf/sosp/pipedream} and DAPPLE~\cite{DBLP:conf/ppopp/dapple} combine pipeline parallelism with data parallelism to improve the efficiency. Unfortunately, existing approaches only support limited parallelism dimensions (i.e., data parallelism and rare model parallelism dimensions) or rely on strong model and hardware configurations (i.e., expert-designed parallelism strategy) and result in sub-optimal performance in practice. 
To the best of our knowledge, there are few prior works considering the automatic parallelism for large-scale Transformers with a complex search space including multiple parallelism dimensions.


In this approach, we propose \name, a novel automatic parallel training system for Transformer models over multiple GPUs. Our target is to integrate data parallelism with a variety of model parallelism dimensions, provide a rarely larger search space (compared with previous approaches), and find the optimal hybrid parallelism strategies in an efficient manner.
However, such an integration brings an explosive growth of the search space and cannot be directly explored as usual. Therefore, we are interested in the following question: \textit{How can we exploit as many parallelism dimensions as possible and efficiently explore the search space in the meanwhile?}

We study five parallelism paradigms, four of which are popular parallelism paradigms in the distributed training of Transformer models, including data parallelism (DP), sharded data parallelism (SDP)~\cite{DBLP:conf/sc/zero}, tensor parallelism (TP), and pipeline parallelism (PP). Besides, we also take into account activation checkpointing (CKPT) as a special parallelism dimension, which distributes the training memory workload to the backward computation through checkpoints.
{These parallelism paradigms} have distinct memory consumption and communication overheads and no single paradigm could beat the others on both sides.
The search space of automatic parallelism should include the arbitrary combinations of them.
Inspired by some key intuitions from our observations and analysis, we first propose a decision-tree structure to decompose the search space and perform pruning to remove the inefficient combinations. 
To determine the final distributed execution plan, we then propose a dynamic programming search algorithm to utilize the optimal substructure property of this problem. 
{
Based on these, we provide \textbf{\name}, which not only targets automatic parallelism for Transformer model training, but also considers the \textbf{B}alancing trade-off between \textbf{M}emory and computation \textbf{W}orkloads across devices.
During the search process, \name provides the required computation and communication costs and memory consumption through a cost estimator.}
It is worth mentioning that the cost estimation in \name considers the GPU performance slowdown from computation and communication overlapping, which has been ignored for a long time in previous approaches.
We provide an implementation of \name over PyTorch. 
Unlike existing toolbox-like systems (e.g., DeepSpeed~\cite{rasley2020deepspeed}, Megatron~\cite{DBLP:conf/sc/megatron}) relying on users' expertise and significant tuning efforts,
\name's automatic parallelism only requires a few lines' modifications on the original training script. Our evaluation selects four representative Transformers, including both NLP (i.e., BERT and T5) and CV (i.e., ViT, Swin Transformer). The experiments show that \name could significantly outperform the four pure parallelisms and existing automatic parallelisms with limited dimensions (i.e., DP+TP and DP+PP) under various device memory budgets.


We summarize our contributions as follows: 

\begin{quote}
1) We enlarge the explored dimension of automatic parallelism for Transformer training to five parallelism dimensions, and introduce a novel decision-tree abstraction to decompose the large search space. \\
2) We design a novel parallelism optimization method to automatically find the most efficient hybrid parallelism strategy based on the estimated costs. \\
3) We consider both memory consumption and computation workload through a bi-objective optimization framework to maximize the hardware utilization during training. \\
4) We build \name system that supports larger models' training and achieves up to $530\%$ and $242\%$ throughput speedups compared to state-of-the-art pure and hybrid parallelism methods respectively.
\end{quote}



\begin{figure}[t]
    \centering
    \includegraphics[width=1.0\linewidth]{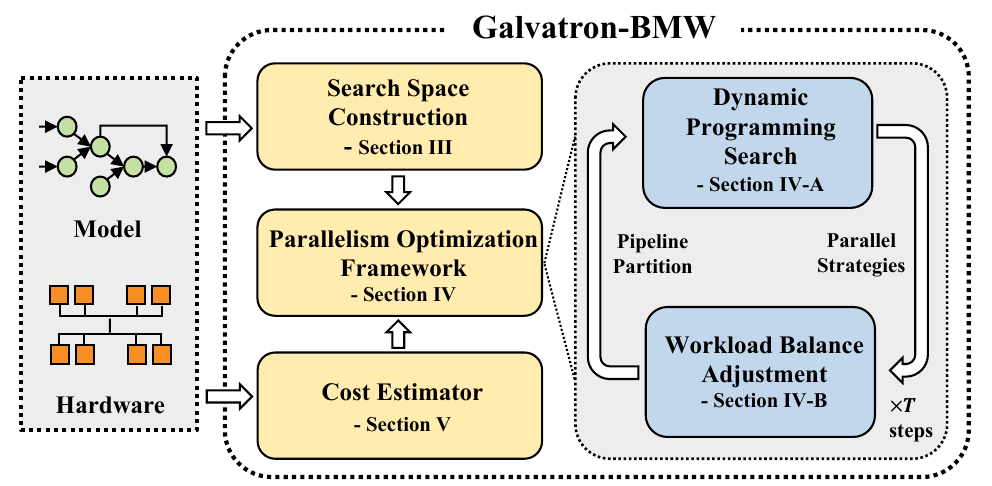}
    \vspace{-5mm}
    \caption{System overview of \name.}
    \label{fig:overview}
    \vspace{-6mm}
\end{figure}

Figure~\ref{fig:overview} shows the system overview of \name, which takes the model and hardware environment as inputs, and comprises three primary modules: search space construction (Section~\ref{sec:search space construction}), parallelism optimization framework (Section~\ref{section:parallel_opt}), and cost estimator (Section~\ref{subsection:cost_model}).
In Section~\ref{sec:search space construction}, we introduce the search space construction of \name with decision-tree-based decomposition. In Section~\ref{section:parallel_opt}, we propose our parallelism optimization framework, which leverages dynamic programming search and workload balance adjustment techniques to iteratively refine the optimal parallelism strategy. In Section~\ref{subsection:cost_model}, we provide a cost estimator to estimate the execution cost and memory consumption efficiently and accurately. We also provide implementation details in Section~\ref{section:imple} and comprehensive experimental results in Section~\ref{section:experiments}.

\section{Preliminary}
\subsection{Transformer Models}
Transformers are first proposed to solve sequence modeling and transduction problems such as language modeling and machine translation~\cite{DBLP:conf/nips/transformer}. 
The self-attention and point-wise feed-forward modules are the basic components in each Transformer layer. 
Most operations are dense algebras like matrix multiplications, resulting in huge computation costs and memory consumption.


\textit{\textbf{Transformers in NLP.}} Different manners of using Transformer layers in NLP incur three mainly Transformer architectures, including encoder-only (for text classification, e.g., BERT and RoBERTa~\cite{DBLP:journals/corr/abs-1907-11692}), decoder-only (for text generation, e.g., GPT-2 and Transformer-XL~\cite{DBLP:conf/acl/DaiYYCLS19}), and encoder-decoder (for sequence-to-sequence tasks, e.g., T5 and BART~\cite{DBLP:conf/acl/LewisLGGMLSZ20}). 
They have similar basic model components and some slight differences on the structures. For example, the decoder has an additional self-attention layer compared to the encoder. What's more, the encoder-decoder architecture combines encoders and decoders symmetrically (i.e., the same number of layers) together. These differences bring some distinct system workload characteristics in both computation and memory.

\textit{\textbf{Transformers in CV.}} Transformers are also becoming increasingly attractive in computer vision areas. Vision Transformer (ViT) first replaces the tokens in languages with patches in images and the patches are fed into the encoder for the image classification task. 
Standard ViTs have a fixed number of patches and the same hidden dimension across different layers. 
Swin Transformer proposes a multi-stage hierarchical architecture with a shifted window-based attention to encode multi-scale patches. However, such multi-scale architectures also cause uneven computation and memory across layers.

\vspace{-4mm}
\subsection{Parallelism in Distributed Training}
\label{subsection:parallel_in_dist_training}

\textit{\textbf{Data parallelism.}} Data parallelism approaches are widely used to scale up the distributed training for large input datasets. It refers to distributing the data samples across multiple workers to compute and synchronize the model updates (e.g., gradients). Each worker should maintain a replica of the model which implies that the model should be fit into the device memory. To alleviate the redundant memory consumption, DeepSpeed ZeRO~\cite{DBLP:conf/sc/zero} (also named FSDP in FairScale~\cite{baines2021fairscale}) has been proposed to partition the model states instead of replicating them. It is similar to model parallelism but still follows the data parallelism computation process except involving additional communications to share the model states.

\textit{\textbf{Model parallelism.}} Model parallelism divides the model into multiple parts and each worker is only responsible for the computation of the partial model. Due to the complexity of DL model architectures, a variety of model parallelism approaches have been proposed with different model partition techniques.
There are mainly two kinds of paradigms commonly used for large-scale Transformers training, including distributed tensor parallelism (TP) and layer-wise pipeline parallelism (PP).
For example, Megatron-LM~\cite{DBLP:conf/sc/megatron} uses TP, which partitions the feed-forward and self-attention modules in Transformers to multiple devices and inserts communication operations (e.g., All-Reduce) to guarantee consistent results.
GPipe~\cite{DBLP:conf/nips/gpipe} first proposes PP, treats each model as a sequence of layers and partitions the model into multiple composite layers across the devices. The workers are organized as a pipeline and transfer intermediate results at the partition boundaries between neighboring partitions. 
{It further splits the mini-batch into smaller micro-batches to reduce the bubbles (i.e., idle time).}
{PipeDream~\cite{DBLP:conf/sosp/pipedream} and 1F1B-Flush~\cite{narayanan2021memory} (also referred to as PipeDream-Flush) are also popular schedules for PP execution. 
In \name, we default to 1F1B-Flush for its advantages of synchronous weight updates and demonstrating the same theoretical bubble rate as GPipe, while being much more memory-efficient. However, 1F1B-Flush causes distinct memory cost across different PP stages, where shallower stages consumes more memory. Such memory workload imbalance can potentially impede system performance, necessitating careful optimization of PP workload balance.}



\textit{\textbf{Activation checkpointing.}} Activation checkpointing (CKPT) is a commonly used technique in large-scale model training, which trades off computation for memory overhead. In the standard training procedure, all intermediate activations computed in forward propagation must be stored for gradient computation, which can be quite memory-intensive for large models. To alleviate the memory overheads, activation checkpointing divides the model into segments and only stores the input activations of each segment during forward pass, discarding the other intermediate results. As a trade-off, these discarded results needs to be recomputed during backward propagation. Compared to other parallelisms that distribute the memory burden across physical devices, activation checkpointing distributes the memory burden across time dimension by delaying the memory consumption of most layers' intermediate activations to the backward pass with lower peak memory pressure.
Therefore, we treat activation checkpointing as a special parallelism dimension that can be combined with other parallelisms naturally.

\textit{\textbf{Automatic parallelism.}} Recent approaches propose to integrate both data and model parallelism and search for better distributed training strategies. For example, FlexFlow, OptCNN, Tofu and TensorOpt consider both tensor parallelism and data parallelism. PipeDream and DAPPLE extend pipeline parallelism and enable data parallelism to replicate each pipeline stage. However, these approaches only explore the combination of data parallelism and at most one single model parallelism dimension. Such limited decision spaces cannot generate efficient enough parallelization plan for many workloads. In fact, industrial companies have taken great efforts to explore better parallelism combinations when training large Transformers on their clusters, such as Turing-NLG~\cite{DBLP:journals/corr/abs-2201-11990} from Microsoft and GPT-3~\cite{DBLP:gpt3} from OpenAI. These evidences suggest that it is necessary to design an automatic parallelization system covering as many parallelism decisions as possible, without relying on strong system tuning experience from human experts.

\vspace{-2mm}
\section{Search Space Construction}
\label{sec:search space construction}
The goal of \name is to automatically search within the composite parallelism space and generate the optimal parallelization plan for the given Transformer model and the distributed environment.
The key challenge comes from the large search space when considering multiple parallelism strategies and making fine-grained decisions for the model parameters. 
In this section, we first introduce the search space of \name and propose our decision-tree-based decomposition to explore the search space more efficiently.


\vspace{-3mm}
\subsection{Search Space Analysis}
\label{sec:search-space}


We first analyze the overhead of training Transformer models, and take an example environment with two GPUs to better illustrate the large search space, optimization target, and necessary constraints. Then we extend the problem to multi-GPU cases.

{
\subsubsection{Overhead Analysis}
A Transformer model can be treated as a sequence of $L$ layers, and each layer $L_i$ contains a set of model parameters $\mathbf{w}_i$, along with their corresponding parameter gradients and optimizer states, and the combination of these three is referred to as the model states $\mathbf{ms}_i$. Due to the back propagation, the forward computation results (i.e., activations) $\mathbf{f}_i$ should be kept inside the device memory. 
Specifically, activations $\mathbf{f}_i$ include boundary activations $\mathbf{bnd}_i$, which are the input tensors of each layer, and the intermediate activations $\mathbf{int}_i$, which is the intermediate results inside of each layer, and it depends on the parallelism strategy that which part of $\mathbf{f}_i$ requires stashing. 
Besides $\mathbf{f}_i$, the calculation of gradients during back propagation may require extra backward activations $\mathbf{b}_i$.}

\begin{figure}[t]
    \centering
    \includegraphics[width=1.0\linewidth]{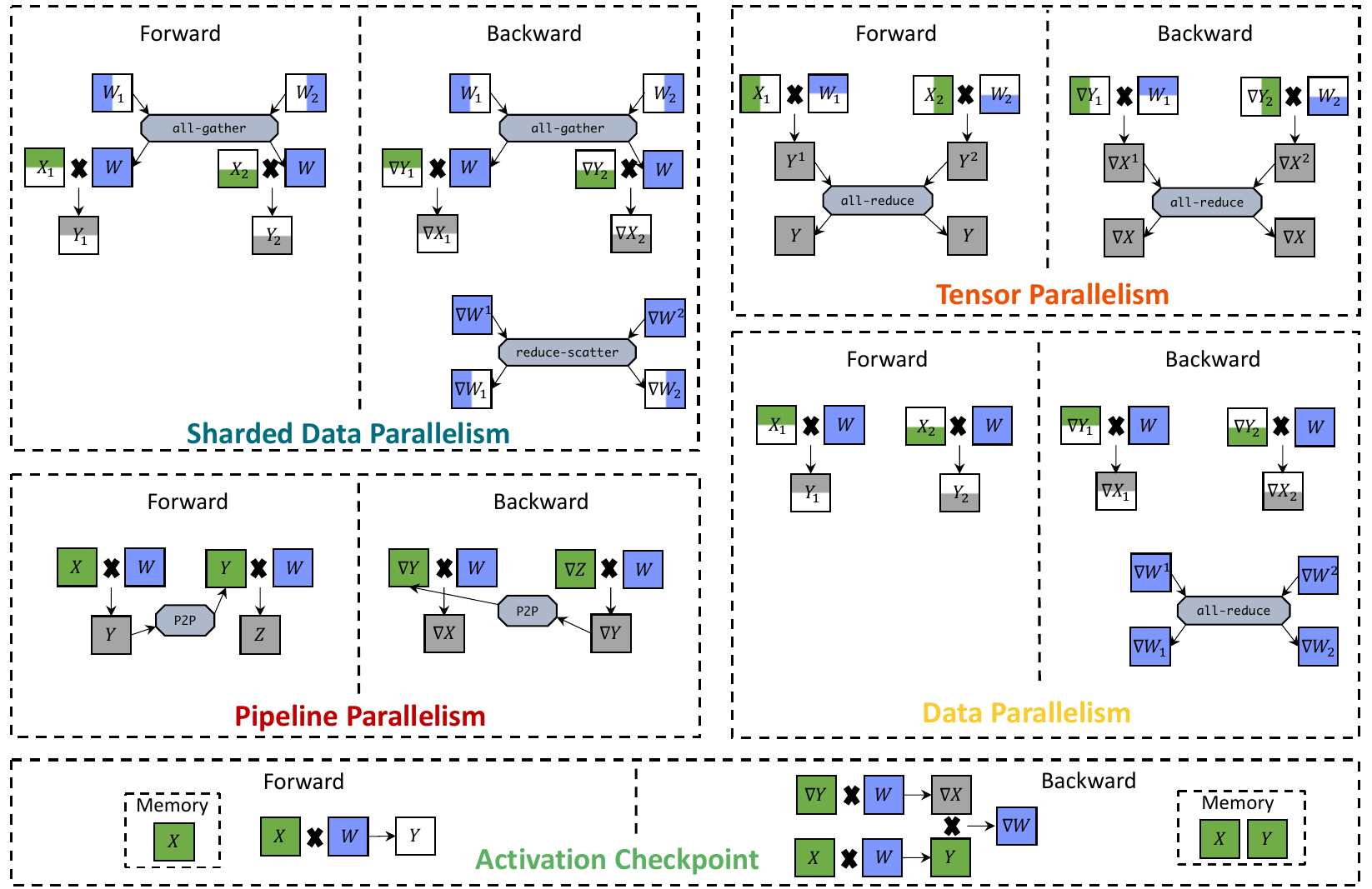}
    \vspace{-6mm}
    \caption{Illustration of different basic parallelisms in \name. We use the green and gray colors to denote the input and output activations for both forward and backward computation. The model parameters and gradients are in blue.}
    \label{fig:parallelism}
    \vspace{-6mm}
\end{figure}

{
The problem is to select the optimal parallelism strategy for each layer individually from a large search space, which is a composition of DP, SDP, PP, TP and CKPT. As illustrated in Figure~\ref{fig:parallelism}, all these parallelism strategies could split the computation workloads into multiple physical devices or distribute memory burden across time dimension,
but they have distinct memory consumption and communication overheads, finally leading to different system efficiency. The overall computation overhead and communication overhead of the Transformer model are the summation of the layer overhead, while the overall memory consumption $E_{all}$ is calculated by formula \ref{eq:E_overall}, where $c(\cdot)$ is the memory cost function.
\vspace{-1mm}
\begin{equation}
    \centering
    \label{eq:E_overall}
    E_{all}=\max_{i=1}^{L} \{\Sigma_{k=1}^{i} c(\mathbf{f}_k) + c(\mathbf{b}_i) +\Sigma_{k=1}^{L} c(\mathbf{ms}_i)\}
    \vspace{-1mm}
\end{equation}
}
{
Specifically, when conducting back propagation of layer $i$, the peak memory cost is the summation cost of activation $\mathbf{f}_1, ..., \mathbf{f}_i$ and $\mathbf{b}_i$, as well as model states $\mathbf{ms}_1, ..., \mathbf{ms}_L$. And $E_{all}$ takes the maximum of the peak memory cost of each layer.
}

\subsubsection{Two-GPU Example}
{
Considering for a single layer $L_i$ in the model, we analyze its costs under different parallelism strategies on two GPUs as follows.
}

\ul{\textit{Data parallelism}}. In DP, each GPU has a model replica and half of the input data samples. Since the size of activations is proportional to the number of data samples, each GPU only needs to store half of the forward activations. After the backward computation, the GPUs should synchronize their gradients (i.e., \texttt{all-reduce}) before updating the model, which has the same size as model parameters.

\ul{\textit{Sharded Data parallelism}}. In SDP, each GPU has half of model parameters and half of the input data samples. However, it requires two times \texttt{all-gather} to collect the sharded model parameters for forward and backward computation and once \texttt{reduce-scatter} to update gradients. Since an \texttt{all-reduce} operation is equivalent to the combination of once \texttt{all-gather} and once \texttt{reduce-scatter}, the communication cost of SDP is 1.5$\times$ larger than DP.


\begin{figure*}[h]
    \centering
    \includegraphics[width=1.0\linewidth]{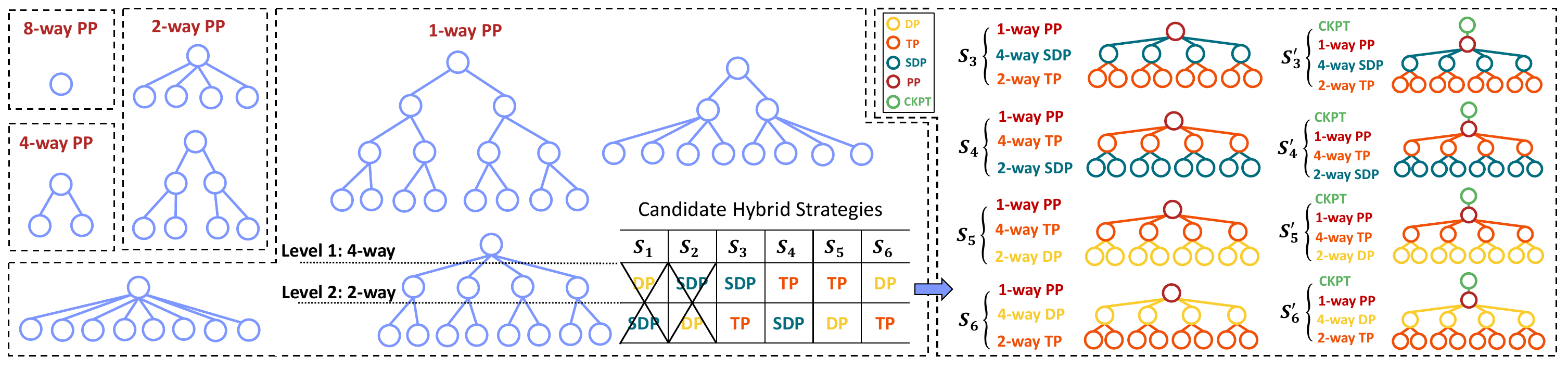}
    \vspace{-6mm}
    \caption{Illustration of the decision trees for 8 GPUs under different PP degrees (i.e., 8/4/2/1). We select one of them to introduce how to use the tree to describe the candidate hybrid parallelism strategies. We remove $S_1$ and $S_2$ as suggested by \textit{Takeaway \#3} and illustrate the other hybrid strategies on the right part. Each decision tree can be decided to apply CKPT ($S_3^{'}$-$S_6^{'}$) or not to apply CKPT ($S_3$-$S_6$). In total, there are 44 candidate hybrid strategies for all trees.}
    \vspace{-6mm}
    \label{fig:tree}
\end{figure*}

\ul{\textit{Pipeline parallelism}}. In PP, the layer $L_i$ could be placed on either GPU 0 or GPU 1, resulting in two possible memory costs: (1, 0) and (0, 1).
The communication cost is mainly determined by whether the neighboring layers are on the same device. 
{In practice, each device may have a sub-sequence of layers and only the activations from the boundary layers should be transferred.}
{The efficiency of PP is also affected by the pipeline bubbles (i.e., idle time), which can be reduced by splitting micro-batches.}
{
Besides, the workload balance also affects its system efficiency, where the model stage with the highest execution cost or the highest memory consumption usually becomes the bottleneck.
}

\ul{\textit{Tensor parallelism}}. In TP, each GPU also has half of model parameters. Unlike SDP, TP allows each device to perform the forward computation (e.g., matrix multiplications and self-attentions) with half model. It requires to synchronize the activations with the \texttt{all-reduce} operations for both forward and backward computation. Due to the intermediate synchronization, TP has some additional replications of the activations.

{
\ul{\textit{Activation checkpointing}}. Unlike other parallelisms that split computation and memory workload over physical devices, CKPT distributes the memory overhead across the time dimension, and brings no communication cost but extra computation cost. It enables the release of intermediate results $\mathbf{int}_i$ during the forward pass and only stores boundary activations $\mathbf{bnd}_i$, which cuts down the forward memory consumption, but requires additional forward computation to recompute $\mathbf{int}_i$ during backward propagation, and brings backward memory consumption as well. Spefically, for layers applying CKPT, $c(\mathbf{f}_i)=c(\mathbf{bnd}_i)$ and $c(\mathbf{b}_i) = c(\mathbf{int}_i)$, while for others not applying CKPT, $c(\mathbf{f}_i)=c(\mathbf{bnd}_i)+c(\mathbf{int}_i)$ and $c(\mathbf{b}_i) = 0$.
}


\subsubsection{Multi-GPU Extension.}
{When extending to multi-GPU, the problem becomes more complicated since even a single layer could have a variety of hybrid parallelism strategies by integrating multiple parallelism paradigms. For example, for two nodes with 4 GPUs in total, it's easy to integrate 2-way TP within a node and 2-way PP across nodes. Alternatively, using 2-way PP within a node and 2-way DP across nodes is also possible. Besides, CKPT can also be combined with other parallelisms, and may bring additional communication overhead
, as it requires an extra forward pass before calculating gradients. For example, when combined with TP, CKPT requires extra \texttt{all-reduce} operations during recomputation.
In the 4-GPU example, it's possible to use 2-way TP and 2-way PP without CKPT, and also valid to use 2-way TP and 2-way PP with CKPT, where CKPT trades-off between memory consumption and time cost. Therefore, CKPT doubles the size of search space further. Moreover, when scaling to 8 GPUs or even more GPUs, there exist hundreds of candidate strategies for a single layer. For a given model, the entire search space is much larger and exponentially growing with the number of layers.}


\vspace{-3mm}
\subsection{Decision-tree-based Search Space Decomposition}
\label{sec:decision-tree}

Considering for such a large search space, it is impossible to brute-force search all the combinations of these parallelism paradigms within a feasible time budget. Therefore, to explore the search space more efficiently, we introduce the following key intuitions from empirical observations or theoretical analysis.

\ul{\textit{Takeaway \#1}}. PP prefers to be applied across device ``islands''. Each island is a set of devices with higher-bandwidth interconnects (e.g., NVLink, PCIe) and should be in charge of a stage in the pipeline. Compared to other parallelisms, PP has much less communication overheads especially for large models. Because each stage typically has multiple layers but only requires to communicate the activations from the boundary layers. It is sensible to perform PP partition first across slower inter-island links (e.g., QPI, Ethernet).

\ul{\textit{Takeaway \#2}}. Suppose the devices are homogeneous, these parallelism strategies prefer to divide the devices into groups with equal size. For example, a 2-way DP on 4 GPUs means two 2-GPU groups, rather than a single GPU and one 3-GPU group. Consequently, the optimal hybrid parallelism strategy on one group should be also consistent with those of the other groups. Note that, it could fail for PP since the model partitions may have different computation operations, resulting in different optimal parallelism strategies.

Based on the above important intuitions, we design a decision-tree to decompose the search space and represent the candidate hybrid parallelism strategies. We next present the details of constructing the decision-tree.

\textit{\textbf{Insights Underpinning Decision-tree.}} We find that most existing automatic parallelism approaches only involve two parallelism dimensions (e.g., OptCNN and FlexFlow), which is easily to enumerate all possible parallelism configurations for a single layer. After involving pipeline parallelism (e.g., PipeDream), they often partition the model into different stages first and each stage is then assigned to a subset of devices. Such kind of observation suggests us to explore the hierarchical search space by utilizing a decision-tree. Another motivation is that we need the tree structure to capture the orders when applying parallelism even inside a stage. Due to the device topology and hierarchical bandwidth, it is necessary to consider the permutations of hybrid strategies since they may have different communication efficiencies.

\textit{\textbf{Decision-tree construction.}}
Given a Transformer model, \name first applies PP to partition the model into multiple stages. In the meanwhile, the devices are also divided into multiple groups with the same size. 
As suggested by Takeaway \#1, it prefers grouping between devices with higher bandwidth.
For an 8-GPU scenario, \name will attempt 1/2/4/8-way PP respectively. 
Suppose the model is partitioned evenly by PP, based on Takeaway \#2, the size of the corresponding device group should be 8/4/2/1 respectively after applying PP, which directly determines the number of leaf nodes in our decision-trees.
As shown in Figure~\ref{fig:tree}, given the number of leaf nodes, there might exist multiple possible tree structures. We define the decision-tree construction rules as follows:

\begin{itemize}
    \item Each decision-tree denotes a sub-search-space and its height represents the number of available parallelism paradigms including DP, TP, PP and SDP.
    \item Any one of DP, TP, PP and SDP cannot be applied repeatedly in different levels of a decision-tree.
    \item The degree of non-leaf nodes should be selected from $\{2,4,8,\cdots\}$.
    \item {Each decision-tree can be decided to apply CKPT ($S_i^{'}$) or not to apply CKPT ($S_i$).}
\end{itemize}


With the above rules, the constructed trees could represent the arbitrary combinations of these parallelisms in a non-overlap manner.
The guidance from \textit{Takeaway \#1} and \#2 significantly helps \name to avoid the unnecessary and inefficient parallelism combinations. For a single layer with 8 GPUs, it produces {68 different hybrid parallelism strategies}
, which reduces the original combinational search space including hundreds of strategies by one order of magnitude. It could be further optimized as follows:

\ul{\textit{Takeaway \#3}}. Using SDP is always better than integrating DP and SDP. We make a comparison with $N$-way DP, $N$-way SDP, and the combination of $N_1$-way DP and $N_2$-way SDP ($N_1\times N_2=N$). First, SDP always has fewer model parameters than DP+SDP since $N_2\leq N$. Second, integrating DP and SDP will lead to two rounds of communication including $2(N_1-1)/N_1$ for $N_1$-way DP and $3(N_2-1)/N_2$ for $N_2$-way SDP. Given $N_1\times N_2=N$, we can prove that the minimum value of its cost is still larger than that of pure SDP. Therefore, we exclude such combinations from our search space.
After applying \textit{Takeaway \#3}, we could further reduce the number of candidate strategies to {44}
for a single layer with 8-GPUs.

\vspace{-2mm}
\section{Parallelism Optimization Framework}
\label{section:parallel_opt}
The target of \name is to generate the optimal hybrid parallelism strategy for the input DL model with the given devices. 

\textit{\textbf{Problem Formulation.}} We define the optimization problem in \name as follows. Given model $M$ (with $L$ layers) and $N$ devices (with memory capacity of $E$), the object is to find the largest throughput $Tpt$ and return the corresponding parallelism strategy, which is made up of the fine-grained layer-level parallelism strategies.
{According to the decision-tree-based decomposition proposed in section~\ref{sec:decision-tree}, PP is first applied to partition the model into multiple stages, and to divide the devices into multiple device groups. Subsequently, the partitioned model stages are assigned to the corresponding device groups, and optimization of other parallelism dimensions is conducted for each model stage. It is evident that pipeline partition plans influence the workload on each device group, thereby affecting the optimization outcome. 
Therefore, we address the optimization problem in two folds:}
\begin{quote}

    \textbf{Question 1:} Given an ideally balanced pipeline partition plan, how to conduct parallelism optimization for each model stage and find the optimal hybrid parallelism strategies?(Section~\ref{subsection:base_opt}, \vldbname-Base)\\
    \textbf{Question 2:} How to find an optimal pipeline partition plan, which balances the pipeline workload (including both memory and computation) and maximizes the system throughput? (Section~\ref{subsection:bi_obj_optimization}, \name)
\end{quote}

\vspace{-4mm}
\subsection{Basic Parallelism Optimization }
\label{subsection:base_opt}
\subsubsection{{Basic Optimization Workflow}}
{We first propose \vldbname-Base to solve \textbf{Question 1} and the optimization algorithm workflow is illustrated in Algorithm~\ref{alg:optimizer}.}
Basically, the system throughput equals to the ratio between the batch size and the iteration time (i.e., per-batch execution time).
Tuning the batch size could lead to distinct memory consumption, computation costs and communication overheads.
Scaling the model training with hybrid parallelism strategies could reduce the memory consumption and enlarge the batch size. But it could also bring significant communication overheads. In other words, the highest training throughput does not have to come with the largest batch size.
{Therefore, in \vldbname-Base, we gradually increase the explored batch size (line 2) and keep tracking the maximum system throughput until exceeding the device memory for all possible parallelism strategies (lines 11-15).}

Given a candidate batch size $B$, \vldbname-Base then utilizes \textit{Takeaway \#1} to apply PP at first. 
We suppose the total number of devices $N$ is the power of two (e.g., 4, 8, 16), which is common in dedicated GPU training clusters.
So we only explore the 2-th powered PP degrees (line 4).
{In default, for each unique PP degree $P$, we assume its pipeline partition plan is given (line 5) and represented as an array $\mathbf{p}$, where each item $\mathbf{p}_i$ denotes the number of model layers for the $i$-th pipeline stage. For example, $\mathbf{p}=[12,12]$ indicates that a 24-layer model is partitioned into two stages with 12 layers each. Note that, the devices are evenly divided into $P$ groups and the model is also partitioned into $P$ stages (line 6), guided by several load balancing factors (e.g., the number of layers/parameters, the maximum memory usage, and the execution time). We leave these further discussion to Section~\ref{subsection:bi_obj_optimization}.
Then we conduct the \texttt{Galvatron\_Search} function to optimize parallel strategies (line 7). In \texttt{Galvatron\_Search} (line 17), we first initialize the micro-batch number $m$ (line 18) and calculate the micro-batch size $B_m$. 
For each model stage, we construct the decision tree that represents the candidate hybrid parallelism strategies composed of DP, SDP, TP, and CKPT.}
{
After obtaining the strategies set $S$, we further determine the parallelization plan for each layer in $M_i$ under the limited device memory budget $E$ with the dynamic programming search algorithm (Section~\ref{sec:dynamic_programming}). 
The search results are used to calculate the whole pipeline execution time (line 27) based on the cost model in Section~\ref{subsection:cost_model}.}

\begin{algorithm}[t]
\scalebox{0.92}{
\begin{minipage}{1.0\textwidth}

\caption{\vldbname-Base Optimization}
\label{alg:optimizer}
\LinesNumbered 
\KwIn{model: $M$, \#devices: $N$, device memory: $E$}
\KwOut{maximum system throughput $Tpt$}
$Tpt \gets 0$\;
\For{Batch size $B \gets 1, 2, ...$}{
    Time costs set $\mathcal{C} \gets \{\}$\;
    \For{PP degree $P \in \{1, 2, 4, 8, ..., N\}$}{
        $\mathbf{p} \gets$ \texttt{PP\_Partition\_Init}$(M,P)$\;
        Model stages $\{M_i\}_{i=1}^{P} \gets \texttt{Model\_Partition}(M, \mathbf{p})$\;
        $C, \mathcal{S} \gets$ \texttt{Galvatron\_Search}{$(\uline{E, \{M_i\}, N, B, P})$}\;
        $\mathcal{C}$\texttt{.append}($C$)\;
    }
    $C_{opt} \gets \min (\mathcal{C})$\;
    \eIf{$C_{opt}$ \textbf{is not} $\infty$}{
        $Tpt \gets \max(B/C_{opt}, Tpt)$\;
    }{
        \textbf{return} $Tpt$ \tcc*{Out-Of-Memory}
    }
}
\SetKwFunction{Fmain}{Galvatron\_Search}
\SetKwProg{Fn}{Function}{:}{}
\Fn{\Fmain{$E, \{M_i\}, N, B, P$}}{
$m \gets \texttt{Init\_Microbatch\_Num}(\{M_i\},B,P)$\;
$B_m \gets B/m$\;
Strategies set $S \gets \texttt{Construct\_Decision\_Tree}(N/P)$\;
Model strategies list $\mathcal{S}$ $\gets []$\;
Model cost $C \gets 0$\;
\For{$i \in \{1, 2, ..., P\}$}{
$C_i, \mathcal{S}_i \gets \texttt{Dynamic\_Programming}(E, M_i, B_m, S)$\;
$\mathcal{S}$\texttt{.extend}($\mathcal{S}_i$)\;
}
$C \gets \texttt{Pipeline\_Cost}(\{C_i\}_{i=1}^{P}, m)$\;
\KwRet{$C, \mathcal{S}$}\;}
\end{minipage}
}
\end{algorithm}

\subsubsection{{Dynamic Programming Search}}
\label{sec:dynamic_programming}
For a given model stage including $L$ layers, we suppose the function $C(L, E)$ represents the total execution time of these $L$ layers under the device memory budget $E$.
{When applying any parallelism strategy $S_j\in S$, we define $c(L, S_j)$ to denote the execution time, $O_f(L, S_j)$ and $O_b(L, S_j)$ to represent the memory consumption of forward activations $\mathbf{f}_L$ and backward activations $\mathbf{b}_L$, and $O_{ms}(L, S_j)$ to denote the memory consumption of model states $\mathbf{ms}_L$.
The overall memory consumption of $L$ layers $E_{all}(L)$ is calculated by Eq.~\ref{eq:E_overall_Sj}. We also calculate the total forward memory consumption of $L$ layers $E_{f}(L)$ by Eq.~\ref{eq:E_f_Sj}, which includes the consumption of forward activations and model states. It is obvious that $E_{f}(L) \leq E_{all}(L)$.
\vspace{-1.5mm}
\begin{equation}
    \centering
    \label{eq:E_overall_Sj}
    \begin{aligned}
    E_{all}(L) =\max_{i=1}^{L} \{ &\Sigma_{k=1}^{i} O_f(k, S_{j_k}) + O_b(i, S_{j_k}) \\
    &\quad +\Sigma_{k=1}^{L} O_{ms}(k, S_{j_k}) \}
    \end{aligned}
\end{equation}
\vspace{-2mm}
\begin{equation}
    \centering
    \label{eq:E_f_Sj}
    \scalebox{1}{
    $E_{f}(L)=\Sigma_{i=1}^{L} \{O_f(i, S_{j_i})+O_{ms}(i, S_{j_i})\}$}
\vspace{-1.5mm}
\end{equation}
We aim to optimize $C(L, E)$ given the memory constraint $E_{all}(L)$ $\leq E$ using dynamic programming search. However, we find that due to the maximum operation in Eq.~\ref{eq:E_overall_Sj}, two memory states need to be stored during state transition, which leads to a quadratic complexity in terms of memory constraint $E$ (refer to Appendix~A1 for details). This is unacceptable in practice.
To ensure the linear complexity with respect to memory, we decouple the forward memory $E_{f}(L)$ from $E_{all}(L)$, and first optimize $C(L, E_{fwd})$ with the forward memory constraint $E_{f}(L) \leq E_{fwd}$, where $E_{fwd} \leq E$, and finally check the validity of the overall memory, i.e., $E_{all}(L)\leq E$.}

{
Firstly, we discuss how to optimize $C(L, E_{fwd})$ with constraint $E_{f}(L) \leq E_{fwd}$ using dynamic programming.
}
Since the problem follows the optimal substructure property (refer to Appendix~A2 for detailed proof), we start with $C(0, \cdot)=0$ and $C(\cdot, 0)=\infty$, then we can derive the following state transition formula:
\vspace{0.5mm}
{
\begin{equation}
\label{eq:search_new}
\resizebox{1.0\linewidth}{!}{
$\begin{aligned}
C(L, E_{fwd}) &= \min_{S_j\in S}\{C(L-1, E_{fwd}-O_f(L, S_j)-O_{ms}(L, S_j)) \\
&\quad + c(L, S_j) + R(L, S_i, S_j)  \}
\end{aligned}$
}
\end{equation}
}
\noindent where 
$R(L, S_i, S_j)$ is the transformation cost between the $L$-th layer applying $S_j$ and its former layer applying $S_i$. If two neighboring layers have different parallelism strategies, the former layer's output should be transformed to the required data layout to facilitate the next layer's parallelism. For example, if the former layer uses the combination between 2-way DP and 2-way TP and the current layer attempts to use 4-way DP, a transformation step is necessary to prepare the full model replica and the $1/4$ forward activation at each device for the current layer. During the state transition process, if the {forward} memory usage exceeds the budget {$E_{fwd}$}, the cost function $C$ should return infinity.

{
Secondly, based on the state transition formula in Eq.~\ref{eq:search_new}, we introduce our overall dynamic programming algorithm (the detailed algorithm is given in Appendix~A3), which ensures $E_{all}(L) \leq E$. Given a device memory budget $E$, we use $E_{fwd}$ ($E_{fwd} \leq E$) as the forward memory budget, and the rest part $E-E_{fwd}$ is spared for the backward peak memory $E_{b}(L)=E_{all}(L)-E_{f}(L)$. To maximize the memory utility, we gradually increase and traverse $E_{fwd}$ to optimize $C(L, E_{fwd})$ using Eq.~\ref{eq:search_new}, and then check the validity of the overall memory with the searched strategies, i.e., $E_{all}(L)\leq E$. 
Finally, we find the largest forward memory budget $E_{fwd}^{opt}$ with its searched strategies satisfying $E_{all}(L)\leq E$, and the optimized throughput $C(L, E_{fwd}^{opt})$ is the final output. 
}



\subsubsection{{Complexity Analysis}}
The proposed dynamic programming search formula in Eq.~\ref{eq:search_new}
has a computation complexity of $\mathcal{O}(LE|S|)$.
As we can see, the size of the single layer's decision space is crucial for the entire complexity and our proposed decision-tree significantly reduces the space and makes it feasible. The number of layers $L$ and the memory budget $E$ also affect the complexity. For extreme cases with thousands of layers or huge memory capacity, we can further reduce the complexity by taking coarse-grained explorations, e.g., fusing multiple layers, using large memory granularity.


\vspace{-2mm}

 \begin{figure}
    \centering
    \includegraphics[width=1.0\linewidth]{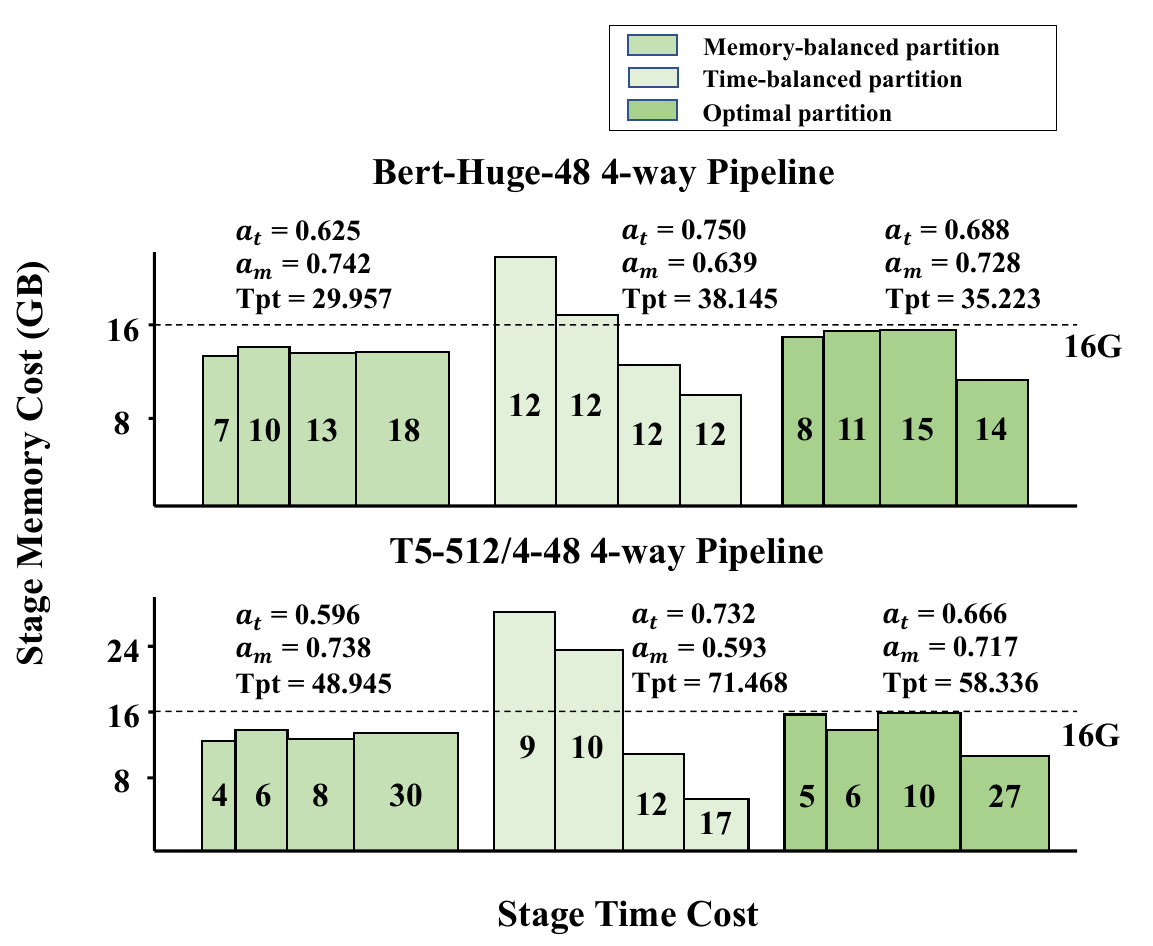}
    \vspace{-5mm}
    \caption{Performance of 4-way 1F1B-Flush pipelines with different partition plans on A100 GPUs. The global batch size is 32 for BERT-Huge-48 and 64 for T5-512/4-48 (see more details of the model in Section~\ref{subsection:exp_setup}), and the micro-batch number is 8.  Bars (from left to right) symbolize pipeline stage 1 through 4: height for memory consumption, width for time cost (normalized), including the number of layers, balance degrees and throughput.}
    \label{fig:observation}
    \vspace{-4mm}
\end{figure}



\vspace{-1mm}
\subsection{{Bi-objective Optimization of Workload Balance}}
\label{subsection:bi_obj_optimization}
{We assume an ideally balanced pipeline partition in \vldbname-Base which may not be true in practice. In fact, perfect workload balance could be difficult to achieve due to several reasons. 
For example, the 1F1B-Flush pipeline scheduling brings distinct memory consumption for different model stages (Section~\ref{subsection:parallel_in_dist_training}). Besides, some model structures (e.g., encoder-decoder) are naturally heterogeneous and hardly to be evenly partitioned based on any single empirical guideline. For instance, the decoder usually has much shorter sequence length than the encoder, leading to workload imbalance of both memory and execution time. Therefore, to address \textbf{Question~2}, we further propose \vldbname-\textbf{BMW} that considers the \textbf{B}alancing trade-off between \textbf{M}emory \textbf{W}orkload and computation workload.}

\subsubsection{{Pipeline Workload Balance}}
{
When training a model $M$ in PP (i.e., no matter using GPipe or 1F1B-flush) with mini-batch size $B$ and memory budget $E$ for each worker, the overall time cost $C(M,B)$ and peak memory cost $O(M,B)$ can be represented as follows:}
\vspace{-3.5mm}
{
\begin{equation}
    \centering
    \label{eq:pipeline_cost_imbalance}
    \scalebox{0.9}{$
        \begin{aligned}
        C(M,B) &= (m - 1) * \max \limits_{i=1}^{P}C(M_i,B_m)+ \sum\limits_{i=1}^{P}C(M_i,B_m) \\[-1.5ex]
        O(M,B) &= \max \limits_{i=1}^{P}O(M_i,B_m) \leq E
        \end{aligned}
$}
\vspace{-1mm}
\end{equation}
}
{
\noindent where $C(M_i,B_m)$ and $O(M_i,B_m)$ denotes the execution time and the memory consumption of the $i^{th}$ model stage $M_i$ under micro-batch size $B_m$ ($B_m = B / m$). As can be seen, the overall pipeline execution time largely depends on the slowest worker, and the memory bottleneck lies in the worker with the heaviest memory workload. 
}
{
To quantify the workload imbalance, we define the following time and memory balance degrees:
}
{
\vspace{-2mm}
\begin{equation}
\centering
\label{eq:balance_deg}
\scalebox{0.9}{$\alpha_t=1-\dfrac{\max\limits_{i=1}^P C(M_i,B_m)}{\sum\limits_{i=1}^P C(M_i,B_m)}, \alpha_m=1-\dfrac{\max\limits_{i=1}^P O(M_i,B_m)}{\sum\limits_{i=1}^P O(M_i,B_m)}$,}
\vspace{-1mm}
\end{equation}
}
{
which satisfy that $0 \leq \alpha_t, \alpha_m \leq 1-\frac{1}{P}$. It's easy to observe that a larger balance degree implies lower total execution time or peak memory usage. Specially, $\alpha_t=1-\frac{1}{P}$ or $\alpha_m=1-\frac{1}{P}$ indicate a perfect workload balance of time or memory.
}

{Figure~\ref{fig:observation} shows two examples of 4-way 1F1B-Flush PP, implementing both memory- and time-balanced partition. 
}
{
The time-balanced pipeline exhibits similar execution costs across all stages, resulting in a quite balanced $\alpha_t$, which is close to $1-\frac{1}{4}=0.75$ for both models.
Nonetheless, its memory consumption is seriously imbalanced due to the influence of 1F1B scheduling, leading to a relatively low $\alpha_m$, where deeper stages consuming considerably less memory than shallower ones.
In contrast, the memory-balanced pipeline demonstrates nearly uniform memory consumption across all stages, resulting in a substantial $\alpha_m$. 
But it allocates more layers to deeper stages for memory balance, increase their execution costs, and reduces $\alpha_t$.
Regarding system efficiency, a greater $\alpha_t$ corresponds to increased throughput for time-balanced pipelines. However, under a limited memory budget (e.g., 16GB), some stages of time-balanced pipelines are prone to memory exhaustion, making it unfeasible.
Inversely, while the memory-balanced pipelines yield lower system throughput, their higher $\alpha_m$ ensures balanced memory budget utilization, thereby facilitating the training process. 
Hence, our objective is to devise a pipeline partition plan that optimizes both $\alpha_m$ and $\alpha_t$, to ensure workload balance and enhance system efficiency.}

{Based on these two balance degrees, we can further define $\mathbf{p}_t$ and $\mathbf{p}_m$ as the extremely balanced time and memory-balanced partition plans, satisfying $\alpha_t(\mathbf{p}_t)\geq \alpha_t(\mathbf{p}), \alpha_m(\mathbf{p}_m)\geq \alpha_m(\mathbf{p})$ for any partition plan $\mathbf{p}$. We can infer that the optimal partition $\mathbf{p}^{*}$ should guarantee:
\vspace{-1mm}
\begin{equation}
\label{eq:inter_state_def}
\begin{aligned}
 \alpha_t(\mathbf{p}_m) \leq &\alpha_t(\mathbf{p}^{*}) \leq \alpha_t(\mathbf{p}_t), \\
 \alpha_m(\mathbf{p}_t) \leq &\alpha_m(\mathbf{p}^{*}) \leq \alpha_m(\mathbf{p}_m).
\end{aligned}
\vspace{-1mm}
\end{equation} 
Because for any partition $\mathbf{p}$, if $\alpha_t(\mathbf{p}) < \alpha_t(\mathbf{p}_m)$, we'd rather choose memory-balanced partition $\mathbf{p}_m$ as it has both better memory and speed balance, and if $\alpha_m(\mathbf{p}) < \alpha_t(\mathbf{p}_t)$, we prefer $\mathbf{p}_t$ for the same reason.
Taking the optimal partition in Figure~\ref{fig:observation} as an example, it effectively mitigates the memory exhaustion observed in time-balanced pipelines (i.e., with median values of $\alpha_t$ and $\alpha_m$) and simultaneously boosts the throughput of memory-balanced pipelines by as much as 19.2\%.
However, it is unlikely to find $\mathbf{p}^{*}$ by optimizing these two co-related balance degrees separately.}

\begin{algorithm}[t]
\scalebox{0.92}{
\begin{minipage}{1.0\textwidth}

\caption{\name Bi-objective Optimization}
\label{alg:bi_objective_opt}
\LinesNumbered 
\SetKwFunction{coop}{Bi-objective Optimization}
\SetKwFunction{adj}{Pipeline Partition Adjustment}
\SetKwProg{Fn}{Function}{:}{}
\KwIn{model: $M$, \#devices: $N$, device memory: $E$, \\
\quad \quad \quad batch size: $B_0$, constant: $r$}
\KwOut{maximum system throughput $Tpt$}
$Tpt \gets 0$\;
Queue $Q \gets \emptyset$\;
\For{Batch size $B \in [B_0-r,B_0+r]$}{
Time costs set $\mathcal{C} \gets \{\}$\;
\For{PP degree $P \in \{2, 4, 8, ..., N$\}}{
$\mathcal{S}_0 \gets $ \texttt{Strategy\_Init$()$}\;
$\mathbf{p}_0 \gets$ \texttt{PP\_Partition\_Init}$(M,P,\mathcal{S}_0)$\;
$Q\texttt{.push}(\mathbf{p}_0)$\;
\While{$Q \neq \emptyset$}{
Pipeline partition $\mathbf{p} \gets Q\texttt{.pop}()$\;
Model stages \\
\hangindent=0.40em
\hangafter=1
$\{M_i\}_{i=1}^{P} \gets$ \texttt{Model\_Partition}$(M, \mathbf{p})$\;
$C, \mathcal{S} \gets$ \texttt{Galvatron\_Search}{$(\uline{E, \{M_i\}, N, B, P})$}\;
$\mathbf{p}' \gets \texttt{PP\_Partition\_Adjust}(\mathbf{p},M,P,\mathcal{S})$\;
\If{$\texttt{Validate}(\mathbf{p}',M,P,\mathcal{S})$}{
$Q\texttt{.push}(\mathbf{p}')$\;}
}
$\mathcal{C}$\texttt{.append}($C$)\;
}
$C_{opt} \gets \min (\mathcal{C})$\;
\eIf{$C_{opt}$ \textbf{is not} $\infty$}{
    $Tpt \gets \max(B/C_{opt}, Tpt)$\;
}{
    \textbf{return} $Tpt$ \tcc*{Out-Of-Memory}
}
}
\end{minipage}
}
\end{algorithm}

\subsubsection{{Bi-objective Optimization Workflow}}
\label{subsection:bi_obj_workflow}
{To achieve balanced workloads across pipeline stages and maximize the system performance, we propose the bi-objective optimization workflow of \name in Algorithm~\ref{alg:bi_objective_opt}.
It basically follows the workflow of Algorithm~\ref{alg:optimizer} and reuses the \texttt{Galvatron\_Search} function. The key difference is that we adjust the pipeline partition iteratively to optimize the workload balance simultaneously. For each iteration, we start from a pipeline partition $\mathbf{p}$ (line 10) that is initialized (line 7) by the memory-balanced partition $\mathbf{p}_m$.
We then conduct \texttt{Galvatron\_Search} to optimize the parallel strategies $\mathcal{S}$ based on $\mathbf{p}$. Since the update of $\mathcal{S}$ might change the pipeline workload balance, we need to adjust the pipeline partition $\mathbf{p}$ correspondingly.}

{Here we provide a heuristic partition adjustment method that greedily cuts down the workload of the slowest pipeline stage, and adjust the pipeline from memory-balanced to time-balanced. In particular, given an input partition plan $\mathbf{p}$, we first find the model stage with the maximum time cost $C_{max}$. Then we move its boundary layer to its adjacent stages and obtains a new partition $\mathbf{p}'$.
The validation function (line 14) prevents the opportunistic adjustment from being harmful to the overall system efficiency by adding limitations on $\mathbf{p}'$, including 1) the time costs of model stages should be no more than the previous maximum cost $C_{max}$, 2) the memory costs of model stages should not exceed memory budget, and 3) the memory costs of model stages should not surpass the maximum stage memory cost under partition $\mathbf{p}_t$.
If all criteria are met, it can be demonstrated that $\mathbf{p}'$ satisfies condition~\ref{eq:inter_state_def} and $\alpha_t(\mathbf{p}') \geq \alpha_t(\mathbf{p})$, indicating $\mathbf{p}'$ is superior to the prior partition $\mathbf{p}$ in time balance (refer to Appendix~B for more details). Then partition $\mathbf{p}'$ is considered a feasible intermediate partition and is pushed to queue $Q$ for subsequent search iterations.
}

\vspace{-2mm}
\section{Cost Estimator}
\label{subsection:cost_model}
\name provides a cost estimator to estimate the computation and communication costs and memory consumption during the optimization process. 
Current methods primarily use \textit{profiling} or \textit{simulating} for estimation.
In \name, we take advantages from both sides and design a cost model to make the estimations cheap, efficient and accurate. 
Specifically, for the memory consumption, we use the shape of a tensor and its data type to calculate its memory. For the computation time, we suppose it could be estimated by the product of the batch size and the per-sample computation time. The latter could be measured by profiling the real layer execution time on a single device. Note that, the Transformers are mainly composed by matrix multiplication operations, so the backward computation is usually twice of the forward computation. For the communication time, we can 
approximate it
by using the size of tensor to be transferred divided by the inter-device connection's bandwidth.

With the above computation and communication cost estimations, $c(l,s)$ (i.e., the cost of a given layer $l$ using a specific parallelism strategy $s\in S$) could be calculated by simulating the execution process. It consists of two steps, e.g., forward and backward computation. 
The simulation for the forward computation is simple and directly sums up the computation and communication costs (i.e., \texttt{all-gather} in SDP and \texttt{all-reduce} in TP). However, during the backward process, DP and SDP enable the computation and communication overlapping, which may bring estimation errors. 
A typical choice is to take the maximum value from the computation and communication costs (e.g., PipeDream~\cite{DBLP:conf/sosp/pipedream}).
Existing automatic parallelism approaches barely notice that modern GPUs simultaneously performing compute kernels and communication primitives (e.g., NCCL~\cite{nccl}) lead to slowdown for both sides.
The performance degradation is mainly from the resource contention of thread warps in GPU streaming multiprocessors.
We find that such contention could slow down the computation and communication by 1.3$\times$ in our evaluations, which is consistent with some recent observations~\cite{DBLP:conf/isca/RashidiD0SSNK21}.
By considering the overlapping slowdown, \name makes more accurate estimations and better optimizations.
{Besides, for layers applying CKPT, we add an extra forward propagation cost on $c(l, s)$ to simulate the recomputation overhead.}
By summing up the layer cost $c(l,s)$, we calculate the cost of pipeline model stage $M_i$, $C(M_i,B_m)$, where $B_m$ is the micro-batch size. Then the overall time cost of the model is estimated according to Eq.~\ref{eq:pipeline_cost_imbalance}, which considers the workload balance. 

\vspace{-4mm}
\section{Implementation}
\label{section:imple}

\name is an automatic parallel training framework especially for Transformer models (open sourced at~\cite{galvatron}), as a part of a novel distributed DL system Hetu~\cite{scis2022hetu,miao2021het,miao2022hetgmp,DBLP:journals/corr/abs-2203-14685,Miao_2022}.
We provide a simple and efficient interface to \name users by making a few lines' modifications on the PyTorch training programs~\cite{DBLP:journals/corr/abs-1912-01703,DBLP:journals/pvldb/LiZVSNLPSVDC20}.

\textit{\textbf{Communication group.}}
We implement all communication primitives with PyTorch NCCL functions.
As \name supports complex hybrid parallelism strategies, there could exist many communication groups among the GPUs in the generated parallelization plan. To avoid the expensive NCCL groups construction overheads, \name maintains a global communication group pool which is created in advance and contains all groups that might be used.

\textit{\textbf{Transformation optimization.}}
We propose an efficient \textit{Slice-Gather} step to perform the transformations automatically between two neighboring layers with different parallelism strategies.
Given the previous layer with strategy A and the current layer with strategy B, the main idea of \textit{Slice-Gather} is to ensure the input activations for the current layer are placed on the devices according to the requirement of strategy B, which has been extensively studied~\cite{xu2021gspmd,DBLP:conf/osdi/ZhengLZZCHWXZXG22}.

\vspace{-2mm}

\section{Experiments}
\label{section:experiments}
\subsection{Experimental Setups}
\label{subsection:exp_setup}


\begin{table}[t]
\centering
\small
\caption{Statistics of Models}
\label{tab:model_config}
\vspace{-2mm}
\scalebox{0.7}{
\begin{tabular}{c|ccccc}
\toprule 
\makecell{Model} & Layer Num  & Hidden Size & Param. Num & Acti. Size/sample\\
\midrule 
BERT-Huge-32 & 32  & 1280 & 672M & 3149.39MB\\
BERT-Huge-48 & 48  & 1280 & 987M & 4657.51MB\\
BERT-xHuge & 128 & 2560 & 10.2B & 24210.05MB \\
ViT-Huge-32 & 32  & 1280 & 632M & 646.5MB\\
ViT-Huge-48 & 48  & 1280 & 947M & 968.59MB\\
ViT-xHuge & 128 & 2560 & 10.1B & 5313.9MB \\
T5-Large-32 & 16 Enc.+16 Dec.  & 1024 & 502M & 4119.66MB\\
T5-Large-48 & 24 Enc.+24 Dec.  & 1024 & 737M & 6107.75MB\\
T5-512/4-32 & 16 Enc.+16 Dec.  & 1024 & 502M & 1777.06MB\\
T5-512/4-48 & 24 Enc.+24 Dec.  & 1024 & 737M & 2473.10MB\\
Swin-Huge-32 & 2/2/26/2  & 320/640/1280/2560 & 701M & 726.59MB\\
Swin-Huge-48 & 2/2/42/2  & 320/640/1280/2560 & 1016M & 1016.8MB\\
GPT3-15B & 48  & 5120 & 15.4B & 32889.04MB\\
GPT3-39B & 48  & 8192 & 39.1B & 58645.34MB\\
GPT3-65B & 80  & 8192 & 64.9B & 97557.98MB\\
\bottomrule
\end{tabular}}
\vspace{-6mm}
\end{table}

{In this section, we compare \name with 4 pure distributed parallelism strategies implemented by the state-of-the-art systems including PyTorch DDP~\cite{DBLP:journals/pvldb/LiZVSNLPSVDC20} for DP, Megatron~\cite{DBLP:conf/sc/megatron} for TP, PyTorch GPipe~\cite{pytorch_gpipe} for PP, and FairScale FSDP~\cite{xu2020automatic} (similar to DeepSpeed ZeRO Stage-3~\cite{DBLP:conf/sc/zero}) for SDP. To help better understand the benefits of each technique in \name, we also provide several auxiliary baselines for further comparisons. Apart from \vldbname-Base, we use \vldbname to represent a variant of \vldbname-Base that disables CKPT. 
Based on that, we further create \vldbname(DP+TP) and \vldbname(DP+PP) to verify the training efficiency of previous automatic parallelism approaches with limited parallelism dimensions (i.e., DP+TP and DP+PP). \vldbname(1F1B+Bi-obj) is based on \vldbname but enables bi-objective optimization (in other words, it can be treated as \name but disables CKPT).
Specially, DeepSpeed 3D is an expert-designed baseline~\cite{deepspeed3d} integrating DP, TP, and PP globally.}
We select NLP models BERT, T5 as well as CV models ViT, Swin Transformer as our experimental models. 
The statistics of models are listed in Table~\ref{tab:model_config}.
We select the Adam optimizer and use the English Wikipedia and ImageNet-1K as input datasets for them respectively.
{
To further verify the performance of \name, we also select an imbalanced model for some experiments, T5-512/4, a variant of T5 model for question answering task SQuAD, where the sequence length is 512 for encoders and 4 for decoders. Due to the shorter sequence length, the decoders have much less memory consumption than the encoder, leading to the memory imbalance.}
Most experiments are evaluated on a single node equipped with 8 Nvidia RTX TITAN 24 GB GPUs using PCIe 3.0. 
For PP, we manually tune the number of micro-batches to minimize the bubbles and estimate its costs. All results are averaged over 100 iterations.

\begin{table*}[t]
\centering
\small
\caption{Comparison with 8 GPUs under different memory constraints. The maximum throughput (samples/s) of each strategy is given, along with the corresponding batch size in the bracket, and OOM denotes Out-Of-Memory.}
\label{tab:overall_results}
\vspace{-2mm}
\scalebox{0.72}{
\begin{tabular}{c|c|cccccccc}
\toprule 
\makecell{Memory} & \makecell{Strategy} & BERT-Huge-32 & BERT-Huge-48 & ViT-Huge-32 & ViT-Huge-48 & T5-Large-32 & T5-Large-48 & Swin-Huge-32 & Swin-Huge-48 \\
\midrule 
8G  & PyTorch DDP (DP)        & OOM  & OOM   & OOM  & OOM & OOM       & OOM   & OOM  & OOM \\ 
    & Megatron (TP)        & OOM  & OOM   & 16.16 (24) & 10.65 (16)  & OOM  & OOM   & 13.47 (24)  & 8.41 (8) \\ 
    & PyTorch GPipe (PP)        & OOM  & OOM   & 20.57 (56) & 16.59 (32) & OOM    & OOM   & 23.61 (40)  & 16.42 (24) \\ 
    & FSDP/ZeRO-3 (SDP)       & 4.65 (8) & OOM   & 33.25 (64) & 15.71 (40) & 5.97 (8)   & OOM   & 24.86 (48)  & 11.92 (32) \\ 
    & DeepSpeed 3D      & 7.79 (8) & OOM   & 30.56 (40) & 14.59 (16) & 8.12 (8)   & OOM    & 26.22 (32)  & 14.27 (16) \\
    & Galvatron (DP+TP)     & OOM  & OOM   & 29.4 (32) & 15.76 (16) & OOM     & OOM   & 26.18 (24) & 14.76 (16) \\ 
    & Galvatron (DP+PP)     & OOM  & OOM   & 31.79 (48) &  {20.93} (24) &  {9.37} (8)   & OOM   & 27.18 (40)  & 17.71 (24) \\ 
    & Galvatron &  {8.16} (8) & OOM   &  {36.58} (56) &  {20.93} (24) &  {9.37} (8)   & OOM   &  {31.33} (48)  &  {21.64} (32) \\ 
    & Galvatron-Base & 17.04 (88) & 9.99 (72) & 64.03 (544) & 35.89 (512) & 11.50 (56) & 6.87 (56) & 34.55 (328) & 25.69 (292) \\
    & Galvatron (1F1B+Bi-obj) & 10.05 (8) & OOM & 41.85 (56) & 23.10 (40) & 9.52 (16) & OOM & 32.95 (56) & 22.25 (48) \\
    & Galvatron-BMW & \textbf{19.06} (184) & \textbf{10.23} (176) & \textbf{71.57} (568) & \textbf{38.75} (544) & \textbf{20.74} (80) & \textbf{12.74} (64) & \textbf{65.34} (360) & \textbf{34.21} (360) \\
\midrule
12G & PyTorch DDP (DP)        & OOM       & OOM   & 14.22 (16)  & OOM & OOM      & OOM   & OOM & OOM \\ 
    & Megatron (TP)        & 5.72 (8)  & OOM  & 16.71 (48)  & 10.99 (32)  & 5.14 (8)  & OOM   & 13.68 (40)  & 9.62 (24) \\ 
    & PyTorch GPipe (PP)        & 9.22 (8)  &  {6.2} (8) & 25.13 (104) & 16.62 (64) & 9.09 (8)     &  {6.83} (8)   & 26.07 (72) & 19.82 (48) \\ 
    & FSDP/ZeRO-3 (SDP)       & 8.91 (16) & 3.15 (8) & 47.41 (112) & 24.24 (72) & 11.26 (16)   & 4.11 (8)   & 37.38 (88) & 21.98 (64) \\
    & DeepSpeed 3D      & 7.79 (8) & 5.35 (8)   & 37.88 (80) & 22.68 (48) & 8.12 (8)   & 5.76 (8)   & 34.14 (72)  & 20.07 (40) \\
    & Galvatron (DP+TP)     & 8.92 (8)  & 5.35 (8)& 42.21 (64) & 17.2 (32) & 9.53 (8)    & OOM   & 37.26 (56)  & 20.18 (32)\\ 
    & Galvatron (DP+PP)     & 9.22 (8)  &  {6.2} (8) &  {50.69} (72) & 24.01 (56) & 11.95 (16)   &  {6.83} (8)  & 35.87 (56) & 21.69 (48)\\ 
    & Galvatron &  {11.39} (16) &  {6.2} (8) &  {50.69} (72) &  {26.63} (72) &  {14.49} (16)    &  {6.83} (8)  &  {41.69} (64) &  {25.42} (64) \\ 
    & Galvatron-Base & 18.66 (72) & 12.14 (136) & 68.46 (568) & 43.78 (536) & 20.66 (80) & 11.14 (72) & 58.56 (360) & 30.70 (328) \\
    & Galvatron (1F1B+Bi-obj) & 15.20 (32) & 6.82 (8) & 54.92 (104) & 31.79 (72) & 15.24 (24) & 9.06 (8) & 57.73 (136) & 27.02 (96) \\
    & Galvatron-BMW & \textbf{24.37} (192) & \textbf{15.97} (192) & \textbf{79.24} (512) & \textbf{46.94} (568) & \textbf{24.07} (96) & \textbf{15.13} (80) & \textbf{76.86} (384) & \textbf{47.78} (376) \\

\midrule
16G & PyTorch DDP (DP)  & 6.39 (8)  & OOM   & 44.40 (64)  & OOM & 7.79 (8)      & OOM   & 28.61 (40) & OOM\\ 
    & Megatron (TP)        & 6.06 (16) & 3.88 (8) & 16.81 (72)  & 11.02 (40)  & 5.14 (8)  & OOM   & 13.83 (56)  & 9.71 (40)\\ 
    & PyTorch GPipe (PP)        & 12.96 (16) & 6.2 (8) & 25.26 (144) & 17.24 (96) & 9.09 (8)    & 6.83 (8)   & 28.23 (104) & 20.11 (64)\\ 
    & FSDP/ZeRO-3 (SDP)       & 12.47 (24) & 6.06 (16) & 59.93 (160) & 32.15 (104) & 14.95 (24)   & 7.16 (16)  & 49.68 (136)  & 26.46 (88)\\ 
    & DeepSpeed 3D      & 8.50 (16) & 5.35 (8)   & 41.67 (128) & 25.45 (72) & 11.52 (16)   & 5.76 (8)   & 37.13 (104)  & 24.12 (64) \\
    & Galvatron (DP+TP)     & 12.59 (16) & 6.19 (8) & 46.02 (88) & 23.97 (48) & 14.52 (16)  & 6.84 (8)  & 44.65 (80)  & 26.51 (48) \\ 
    & Galvatron (DP+PP)     & 13.00 (16) & 6.2  (8) & 54.05 (120) & 28.01 (56) & 14.64 (16)   & 6.83 (8)  & 44.15 (96) &  25.82 (56)\\ 
    & Galvatron &  {15.05} (24) &  {7.46} (16) &  {63.25} (160) &  {35.74} (104) &  {16.50} (24)     &  {8.36} (16)  &  {54.06} (136) &  {29.21} (72) \\ 
    & Galvatron-Base & 21.70 (128) & 12.92 (96) & 75.98 (744) & 45.75 (656) & 23.67 (88) & 15.69 (80) & 68.98 (264) & 41.15 (360) \\
    & Galvatron (1F1B+Bi-obj) & 16.61 (32) & 10.03 (32) & 88.86 (480) & 39.39 (104) & 16.72 (24) & 9.69 (16) & 73.20 (392) & 39.34 (136) \\
    & Galvatron-BMW & \textbf{27.90} (224) & \textbf{17.37} (208) & \textbf{89.22} (520) & \textbf{50.34} (480) & \textbf{26.75} (128) & \textbf{17.05} (112) & \textbf{84.52} (428) & \textbf{52.78} (392) \\
\midrule
20G & PyTorch DDP (DP)        & 11.57 (16) & OOM   & 61.54 (112)  & 17.02 (32) & 14.3 (16)   & 5.43 (8)  & 42.82 (80)  & 11.8 (24)\\ 
    & Megatron (TP)       & 6.06 (16) & 3.88 (8) & 16.11 (88)  & 11.02 (56) & 5.47 (16)     & 3.55 (8)  & 13.84 (72) & 9.79 (48)\\ 
    & PyTorch GPipe (PP)        & 13.52 (24) & 7.05 (16) & 28.64 (192) & 17.96 (128) & 9.53 (16)    & 8.13 (16)  & 29.75 (128)  & 20.73 (88)\\ 
    & FSDP/ZeRO-3 (SDP)       & 17.06 (40) & 7.8 (24) & 63.75 (216) &  38.29 (136) & 17.93 (32)     & 7.16 (16)  & 55.22 (176) & 32.63 (120)\\
    & DeepSpeed 3D      & 8.50 (16) & 5.35 (8)   & 43.36 (168) & 27.82 (104) & 13.14 (24)   & 7.96 (16)   & 40.60 (136)  & 26.09 (96) \\
    & Galvatron (DP+TP)     & 14.65 (24) & 8.05 (16) & 61.54 (112) & 28.69 (72) & 15.35 (24)   & 6.84 (8)  & 54.87 (104)  & 30.59 (72)\\ 
    & Galvatron (DP+PP)     & 15.52 (24) & 8.11 (16) & 61.54 (112) & 34.88 (96) & 17.27 (24)   &  {10.33} (16)  & 50.19 (136)  & 31.62 (80)\\ 
    & Galvatron &  {18.21} (40) &  {8.95} (24) &  {70.5} (152) &  {41.2} (136) &   {18.64} (32)      &  {10.33} (16) &  {60.06} (144)  &  {37.75} (120) \\ 
    & Galvatron-Base & 22.57 (160) & 14.90 (136) & 77.96 (1232) & 52.06 (800) & 28.06 (96) & 16.33 (96) & 73.26 (392) & 47.43 (520) \\
    & Galvatron (1F1B+Bi-obj) & 22.63 (64) & 11.16 (32) & 91.65 (568) & 51.07 (480) & 26.81 (64) & 11.45 (24) & 76.10 (436) & 51.20 (360) \\
    & Galvatron-BMW & \textbf{29.08} (256) & \textbf{18.30} (232) & \textbf{95.59} (592) & \textbf{52.06} (800) & \textbf{29.28} (152) & \textbf{18.09} (128) & \textbf{87.23} (512) & \textbf{57.23} (456) \\
\bottomrule
\end{tabular}}
\vspace{-4.5mm}
\end{table*}

\vspace{-2mm}
\subsection{End-to-End Comparison}
\autoref{tab:overall_results} shows the overall system throughput results of different models under different strategies with different memory constraints, along with the corresponding batch size.
As we can see, under different model scales and memory budgets, \name always outperforms all baselines in multiple regards. 
{For instance, on ViT, \name promotes the overall system throughput by up to 493\% compared with pure parallelism strategies, and achieves a maximum of 173\% acceleration compared with hybrid strategies, including automatic parallelism with limited dimensions and expert-designed 3D parallelism.}
{Similarly, on the other three models, \name achieves a maximum of 407\%-530\% and 150\%-242\% compared with single and hybrid strategies respectively.}

It is worthy to note that even for models that can fit into GPU memory, TP, PP, or SDP, which incur additional communication overheads compared to DP, may still outperform DP in terms of overall system throughput. 
This phenomenon is not surprising for two reasons.
First, DP has to replicate the entire model, limiting the achievable batch size, which may lead to under-utilization of GPUs.
Second, even with the same batch size, the communication overheads in model parallelism mainly come from activations, which can sometimes be smaller than gradients in DP.
By balancing multiple parallelism dimensions, \name achieves an optimal trade-off between memory, computation, and communication costs, resulting in larger batch sizes and higher throughput.

%

{Then, we look carefully into the effectiveness of each technique in \name. In comparison to \vldbname, we observe that by integrating CKPT into search space, \vldbname-Base amplifies the overall system throughput by up to an impressive 109\%. Such enhancement is attributed to CKPT's memory efficiency, which facilitates \vldbname-Base to achieve larger training batch size (e.g., a batch size of 160 for BERT-Huge-32 under 20 GB memory constraint), thereby optimizing throughput. Furthermore, \vldbname(1F1B+Bi-obj), which enables bi-objective optimization and strikes a balance between memory and computational workload, bolsters the system throughput by up to 44\% compared to \vldbname. Benefiting from these two techniques, \name manages to achieve state-of-the-art performance across all models.}


We can also find that different models may have different preferences on the parallelism strategies. For example, under different memory budgets, BERT almost always prefers DP+PP among all baselines. Similar observations could be also found on some cases of T5. For ViT and Swin, the preferences change to SDP when increasing the memory budgets. The reason mainly comes from that NLP models have larger activation while CV models have larger model parameters, thus the latter could benefit more from sharding the model parameters across the GPUs. Here DeepSpeed 3D uses an officially suggested strategy~\cite{deepspeed3d} combining 2-way DP/TP/PP together. Such a fixed strategy outperforms three pure parallelisms but fails to beat SDP in most cases.
Another interesting finding is that the hybrid parallelisms like DP+TP and DP+PP may perform worse than pure SDP (e.g., ViT-Huge-32 with 8G, Swin-Huge-32 with 16G).
It further indicates that existing automatic parallelism approaches focusing on limited model parallelism dimensions are suffering from these limitations.

\begin{table*}[t]
\centering
\small
\caption{Comparison with 16 GPUs.}
\vspace{-2mm}
\label{tab:16gpus_results}
\scalebox{0.72}{
{\begin{tabular}{c|c|cccccc}
\toprule 
\makecell{Memory} & \makecell{Strategy} & BERT-Huge-32 & BERT-Huge-48 & ViT-Huge-32 & ViT-Huge-48 & T5-512/4-32 & T5-512/4-48\\
\midrule 
8G  & PyTorch DDP (DP)          & OOM         & OOM             & OOM  & OOM & OOM & OOM\\ 
Low-perf & Megatron (TP)          & OOM         & OOM             & 16.86 (32) & 10.86 (16) & 11.38 (16) & OOM\\ 
Cluster & PyTorch GPipe (PP)        & 13.79 (16)  & 5.88 (8)        & 50.70 (128) & 27.96 (80) & 21.45 (32) & 14.95 (32)\\ 
    & FSDP/ZeRO-3 (SDP)         & 8.95 (16)   & 6.12 (16)       & 69.48 (128) & 34.92 (96) & 31.63 (48) & 15.94 (32)\\ 
    & DeepSpeed 3D              &   {15.24} (16) & 6.43 (8) & 57.14 (64) & 29.92 (40) & 27.97 (32) & 14.03 (16) \\ 
    & Galvatron (DP+TP)         & OOM  & OOM                    & 54.43 (64)  & 24.56 (32) & 22.37 (16) & 15.21 (16)\\ 
    & Galvatron (DP+PP)         & 13.91 (16)  & 5.88 (8)            & 68.56 (128) & 35.02 (72) & 33.63 (40) & 18.99 (32)\\ 
    & Galvatron          &   {15.24} (16) &   {8.43} (16)   &   {76.74} (128) &   {38.32} (88) &   {36.14} (48) &   {21.25} (32)\\ 
    & Galvatron (1F1B+Bi-obj)   & 25.02 (88)  & 12.00 (64)   & 120.90 (544) & 66.77 (384) & 54.71 (72) & 23.07 (48) \\ 
    & Galvatron-Base          & 38.41 (160)  & 15.06 (128)   & 111.21 (672) & 62.84 (648) & 63.84 (192) & 34.38 (208)\\ 
    & Galvatron-BMW           & \textbf{42.04} (256) & \textbf{20.36} (224)  & \textbf{139.51} (928) & \textbf{78.32} (784) & \textbf{94.03} (256) & \textbf{56.49} (256)\\ 
    
\midrule 
16G & PyTorch DDP (DP)        & 12.14 (16)  & OOM   & 88.06 (128)  & OOM & 43.05 (48) & 10.92 (16)\\ 
Low-perf & Megatron (TP)        & 6.12 (16) & 4.23 (16) & 17.11 (64)  & 11.26 (48)  & 12.53 (32) & 8.00 (16)\\ 
Cluster & PyTorch GPipe (PP)        & 23.29 (40) & 12.92 (24) & 69.72 (320) & 50.23 (208) & 36.09 (80) & 26.11 (80)\\ 
    & FSDP/ZeRO-3 (SDP)       & 30.37 (64) & 11.74 (32) & 123.95 (320) & 61.49 (224) & 68.92 (112) & 36.33 (80)\\ 
    & DeepSpeed 3D           & 23.92 (48) & 13.03 (24) & 91.56 (256) & 53.81 (152) & 49.23 (88) & 28.45 (56)\\ 
    & Galvatron (DP+TP)     & 23.01 (32) & 10.50 (16) & 99.22 (160) & 49.82 (96) &   {73.82} (80) & 34.71 (48)\\ 
    & Galvatron (DP+PP)     & 23.73 (40) & 13.12  (40) & 115.88 (224) & 61.38 (208) & 69.42 (96) & 32.76 (56)\\ 
    & Galvatron &   {32.67} (64) &   {14.74} (40) &   {131.15} (320) &   {72.74} (208) &   {73.82} (80) &   {40.10} (64) \\ 
    & Galvatron (1F1B+Bi-obj)     & 49.65 (232)  & 22.30 (104)   & 167.48 (832) & 100.26 (512) & 92.90 (320) & 68.94 (288) \\ 
    & Galvatron-Base     & 48.66 (368)  & 25.11 (200)  & 163.92 (1376) & 91.85 (904) & 87.58 (384) & 55.12 (288) \\ 
    & Galvatron-BMW  & \textbf{55.57} (384) & \textbf{33.14} (320)   & \textbf{179.89} (960) & \textbf{114.08} (1136) & \textbf{117.71} (400) & \textbf{80.53} (448) \\ 

\midrule 
8G  & PyTorch DDP (DP)          & OOM & OOM & OOM & OOM & OOM & OOM \\ 
High-perf & Megatron (TP)            & OOM & OOM & 53.43 (32) & 31.92 (16) & 30.75 (16) & OOM \\ 
Cluster & PyTorch GPipe (PP)    & 57.22 (16) & 14.74 (8) & 233.45 (128) & 176.60 (80) & 84.23 (32) & 55.81 (32) \\ 
    & FSDP/ZeRO-3 (SDP)         & 27.75 (16) & 17.80 (16) & 204.73 (128) & 119.91 (96) & 107.13 (48) & 47.11 (32) \\  
    & DeepSpeed 3D              & 85.38 (16) & 42.39 (8) & 292.64 (64) & 185.23 (40) & 124.71 (32) & 70.36 (16) \\ 
    & Galvatron (DP+TP)         & OOM & OOM & 126.05 (48) & 38.65 (32) & 48.02 (16) & 32.69 (16) \\ 
    & Galvatron (DP+PP)         & 74.80 (16) & 35.42 (8) & 412.70 (104) & 244.97 (80) & 142.34 (40) & 75.05 (32) \\ 
    & Galvatron          &   {99.17} (24) &   {53.38} (16)  &   {453.87} (152) &   {277.94} (96) &   {152.71} (48) &   {83.86} (32)\\ 
    & Galvatron (1F1B+Bi-obj)   & 162.52 (96) & 81.04 (64) & 572.05 (576) & 407.64 (384) & 216.47 (128) & 93.53 (48) \\ 
    & Galvatron-Base          & 140.21 (144) & 94.33 (136) & 576.94 (832) & 369.09 (704) & 227.40 (288) & 151.96 (248) \\ 
    & Galvatron-BMW           & \textbf{202.34} (296) & \textbf{132.03} (288)  & \textbf{697.34} (1152) & \textbf{457.74} (1056) & \textbf{348.29} (416) & \textbf{230.74} (384)\\ 
    
\midrule 
16G  & PyTorch DDP (DP)         & 34.95 (16) & OOM & 303.26 (128) & OOM & 126.02 (48) & 28.15 (16) \\ 
High-perf & Megatron (TP)            & 18.87 (16) & 12.55 (16) & 55.68 (64) & 35.05 (48) & 35.18 (32) & 21.97 (16) \\ 
Cluster & PyTorch GPipe (PP)    & 91.10 (40) & 27.42 (24) & 323.95 (320) & 254.12 (208) & 154.53 (80) & 98.55 (80) \\ 
    & FSDP/ZeRO-3 (SDP)         & 113.98 (64) & 38.89 (32) & 514.80 (320) & 274.47 (224) & 212.35 (112) & 107.24 (80) \\  
    & DeepSpeed 3D              & 133.67 (48) & 82.25 (32) & 506.17 (256) & 322.73 (152) & 188.74 (88) & 99.94 (56) \\ 
    & Galvatron (DP+TP)         & 72.37 (32) &  25.65 (16) & 482.42 (208) & 165.92 (96) & 200.28 (80) & 95.48 (48) \\ 
    & Galvatron (DP+PP)         & 135.48 (48) & 74.95 (32) & 611.15 (248) & 346.13 (128) & 230.16 (96) & 115.55 (56) \\ 
    & Galvatron          &   {147.26} (64) &   {95.90} (40)  &   {641.40} (352) &   {398.28} (232) &   {238.56} (96) &   {127.34} (72)\\ 
    & Galvatron (1F1B+Bi-obj)   & 226.73 (192) & 144.59 (144) & 797.16 (920) & 488.44 (896) & 349.72 (192) & 207.08 (192) \\ 
    & Galvatron-Base          & 168.08 (376) & 113.63 (344) & 727.98 (1344) & 440.82 (1152) & 323.79 (440) & 213.65 (400) \\ 
    & Galvatron-BMW           & \textbf{243.55} (416) & \textbf{161.07} (320)  & \textbf{871.26} (1344) & \textbf{533.46} (1088) & \textbf{409.47} (512) & \textbf{265.37} (480)\\ 
    
\bottomrule
\end{tabular}}
}
\vspace{-4.5mm}
\end{table*}




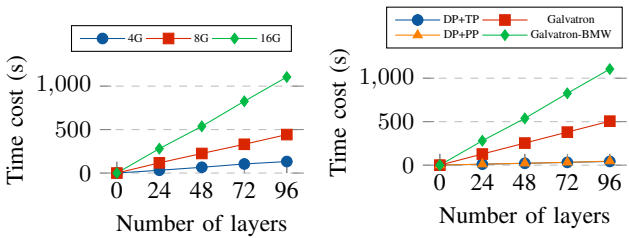
\begin{figure}
\begin{minipage}[h]{0.23\textwidth}
\begin{tikzpicture}
\small
\begin{axis}[
        legend style={at={(0.4,1.4)},anchor=north, nodes={scale=0.5, transform shape}},
        xlabel={Number of layers},
        ylabel={Time cost (s)}, 
        symbolic x coords={0, 24, 48, 72, 96},
        xticklabels={},
        extra x ticks={0, 24, 48, 72, 96},
        extra x tick labels={0, 24, 48, 72, 96},
        ymajorgrids,
        scale only axis,
        width=0.65\linewidth,
        height=40pt,
        ymin=0,
        ylabel shift=-3pt,
        ylabel near ticks,
        xlabel near ticks,
        xtick pos=bottom,
        y axis line style={opacity=0},
        grid style=dashed,
        legend style={align=left},
        legend columns=-1,
    ]
    \addplot+[sharp plot, mark=oplus*, color={rgb:red,12;green,93;blue,165}, mark options={draw=none}] coordinates {
        (0, 0) (24, 32.18313003) (48, 66.27999425) (72, 104.4877009) (96, 132.5865083)
    };
    \addplot+[sharp plot, mark=square*, color={rgb:red,255;green,44;blue,0}, mark options={draw=none}] coordinates {
        (0, 0) (24,  117.553076) (48, 225.1840918) (72, 330.9772432) (96, 442.6173906)
    };
    \addplot+[sharp plot, mark=diamond*, color={rgb:red,0;green,185;blue,69}, mark options={draw=none}] coordinates {
        (0, 0) (24, 280.5868008) (48, 537.9133041) (72, 825.8608432) (96, 1104.916837)
    };
    \legend{4G, 8G, 16G}
\end{axis}
\end{tikzpicture}
\vspace{-6mm}
\subcaption{Different memory budgets}
\end{minipage}
\begin{minipage}[h]{0.23\textwidth}
\begin{tikzpicture}
\small
\begin{axis}[
        legend style={at={(0.4,1.5)},anchor=north, nodes={scale=0.5, transform shape}, legend columns=2},
        xlabel={Number of layers},
        ylabel={Time cost (s)}, 
        symbolic x coords={0, 24, 48, 72, 96},
        xticklabels={},
        extra x ticks={0, 24, 48, 72, 96},
        extra x tick labels={0, 24, 48, 72, 96},
        ymajorgrids,
        scale only axis,
        width=0.65\linewidth,
        height=40pt,
        ymin=0,
        ylabel shift=-3pt,
        ylabel near ticks,
        xlabel near ticks,
        xtick pos=bottom,
        y axis line style={opacity=0},
        grid style=dashed,
    ]
\addplot+[sharp plot, mark=oplus*, color={rgb:red,12;green,93;blue,165}, mark options={draw=none}] coordinates {
    (0, 0) (24, 10.81789231) (48, 20.97259283) (72, 31.61702681) (96, 42.26146078)
};

\addplot+[sharp plot, mark=square*, color={rgb:red,255;green,44;blue,0}, mark options={draw=none}] coordinates {
    (0, 0) (24, 127.7061207) (48, 253.7028897) (72, 379.6996586) (96, 505.6964276)
};

\addplot+[sharp plot, mark=triangle*, color=orange, mark options={draw=none}] coordinates {
    (0, 0) (24, 10.81789231) (48, 21.54467845) (72, 32.13630366) (96, 46.23788142)
};

\addplot+[sharp plot, mark=diamond*, color={rgb:red,0;green,185;blue,69}, mark options={draw=none}] coordinates {
    (0, 0) (24, 280.5868008) (48, 537.9133041) (72, 825.8608432) (96, 1104.916837)
};

\legend{DP+TP,  Galvatron, DP+PP, \name}
\end{axis}
\end{tikzpicture}
\vspace{-6mm}
\subcaption{\name vs. Galvatron vs. DP+TP vs. DP+PP}
\end{minipage}
\vspace{-2mm}
\caption{Search time costs with different numbers of layers.}\label{fig:Search_time}
\vspace{-5mm}
\end{figure}
\vspace{-4mm}
\subsection{Optimization Efficiency}
\vspace{-0.5mm}
The efficiency of our dynamic programming search algorithm varies according to different number of model layers, overall strategies and memory constraints. As shown in \autoref{fig:Search_time} (a), when the number of model layers and memory limit increase linearly, the search time of our algorithm increases linearly as excepted, only hundreds of seconds are required to generate the optimal execution plan, which is acceptable and negligible relative to the extremely long model training time. \autoref{fig:Search_time} (b) demonstrates the impact of total parallelism dimensions on the search time, both DP+TP and DP+PP have a total of 4 alternate strategies on 8 GPUs, while \name and \vldbname has 44 and 22 overall candidates. In this case, the search time of DP+TP and DP+PP is consistent and much less than that of \vldbname and \name.

\vspace{-4.5mm}
\subsection{Scalability Study}
\vspace{-0.5mm}
We conduct further comparisons on large clusters. We first extend our experiments to 16 Nvidia RTX TITAN GPUs over two servers connected by 100 Gb InfiniBand network {(referred to as low-performance cluster), as well as 16 Nvidia A100 GPUs with NVLink over two servers connected by 100Gb InfiniBand network (referred to as high-performance cluster).}
Table~\ref{tab:16gpus_results} illustrates the results on BERT, ViT and T5-512/4 models. Not surprisingly, \name achieves the best performance with different memory budgets {on both two clusters}. {On low-performance cluster}, compared with the results on 8 GPUs, \name
and the hybrid parallelism methods 
could obtain more than 2$\times$ speedups for many cases.

\begin{table}[t]
\centering
\small
\caption{Comparison with 64 GPUs.}
\vspace{-2mm}
\label{tab:64gpus_results}
\scalebox{0.8}{
{\begin{tabular}{c|c|cc}
\toprule 
\makecell{Memory} & \makecell{Strategy} & BERT-xHuge & ViT-xHuge \\
\midrule 
16G  & PyTorch DDP (DP)        & OOM  & OOM  \\ 
High-perf  & Megatron (TP)        & 0.68 (3)  & 1.94 (12)  \\ 
Cluster  & PyTorch GPipe (PP)        & 9.74 (16)  & 61.95 (96)   \\ 
    & FSDP/ZeRO-3 (SDP)       & OOM & OOM   \\ 
    & DeepSpeed 3D          & 8.44 (16) & 64.91 (96) \\ 
    & Galvatron (DP+TP)     & 1.73 (4)  & 5.07 (2)   \\ 
    & Galvatron (DP+PP)     & 9.74 (16)  & 64.83 (104) \\ 
    & Galvatron &   {13.77} (24) &   {68.35} (136)  \\ 
    & Galvatron (1F1B+Bi-obj)   & 19.43 (24) &  109.81 (256) \\ 
    & Galvatron-Base          & 25.30 (128) &  130.88 (768) \\ 
    & Galvatron-BMW           & \textbf{37.09} (376) & \textbf{209.48} (1392) \\ 
\midrule 
32G  & PyTorch DDP (DP)        & OOM  & OOM  \\ 
High-perf  & Megatron (TP)        & 0.77 (7)  & 2.11 (28)  \\ 
Cluster  & PyTorch GPipe (PP)        & 21.38 (48)  & 94.84 (288)   \\ 
    & FSDP/ZeRO-3 (SDP)       & OOM & OOM    \\ 
    & DeepSpeed 3D & 21.28 (40) & 91.19 (256) \\ 
    & Galvatron (DP+TP)     & 1.73 (4) & 5.51 (68)   \\ 
    & Galvatron (DP+PP)     & 23.64 (48)  & 110.98 (232) \\ 
    & Galvatron      &   {27.49} (64) &   {114.55} (328)  \\ 
    & Galvatron (1F1B+Bi-obj)   & 41.11 (80) &  180.08 (640) \\ 
    & Galvatron-Base          & 40.17 (184) &  195.40 (1024) \\ 
    & Galvatron-BMW           & \textbf{60.04} (640) & \textbf{330.30} (1744) \\ 
\bottomrule
\end{tabular}}
}
\vspace{-6mm}
\end{table}

\begin{figure*}[t]
    \centering
    \includegraphics[width=1.0\linewidth]{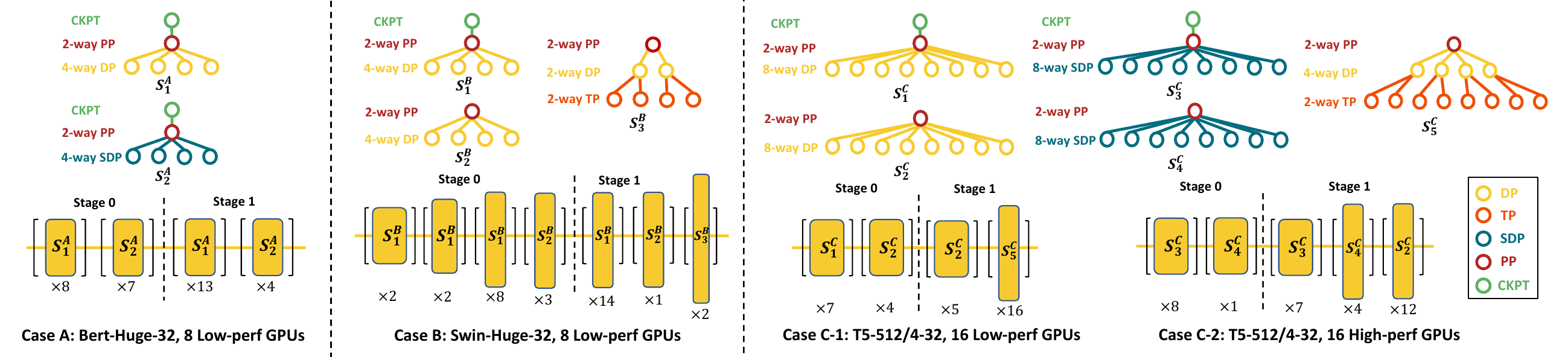}
    \caption{Examples of the optimal parallelism plans given by \name. Case A and B is for BERT-Huge-32 and Swin-Huge-32 under 8 GB memory budgets on 8 low-performance GPUs, and case C is for T5-512/4-32 under 8GB memory budgets on 16 low-performance GPUs and 16 high-performance GPUs.
    Each yellow rectangle denotes an encoder layer, and its height and width represent parameter size and activation size respectively. The number $\times N$ under the rectangle means applying an strategy for consecutive $N$ layers.}
    \label{fig:opt_strategy}
\end{figure*}

\begin{table*}
\centering
\small
\vspace{-1mm}
\caption{Ablation study on bi-objective optimization of workload balance on high-performance cluster (16 A100). The performance of bi-objective optimization (1F1B+Bi-obj) is compared with that of memory-balanced partition (1F1B+Mem) and time-balanced partition (1F1B+Time). The pipeline partition $\mathbf{p}$ is shown next to the system throughput and training batch size.}
\vspace{-2mm}
\label{tab:bi_obj_results}
\resizebox{\textwidth}{!}{%
\begin{tabular}{c|c|cc|cc|cc|cc}
\toprule 
\makecell{Memory} & \makecell{Strategy} & \multicolumn{2}{c|}{BERT-Huge-32} & \multicolumn{2}{c|}{BERT-Huge-48} & \multicolumn{2}{c|}{T5-512/4-32} & \multicolumn{2}{c}{T5-512/4-48} \\
\midrule 
8G  & Galvatron (1F1B+Mem)   & 139.27 (96) & [11,21] & 74.13 (64) & [6,9,13,20] & 198.23 (128) & [6,26] & 87.19 (48) & [9,39] \\ 
High-perf & Galvatron (1F1B+Time)   & 104.17 (24) & [16,16] & 53.38 (16) & [24,24] & 161.52 (48) & [14,18] & 82.45 (24) & [10,10,13,15] \\ 
Cluster & Galvatron (1F1B+Bi-obj)  & \textbf{162.52} (96) & [14,18] & \textbf{81.04} (64) & [7,9,13,19] & \textbf{216.47} (128) & [8,24] & \textbf{93.53} (48) & [12,36] \\ 
\midrule 
16G  & Galvatron (1F1B+Mem)   & 201.54 (376) & [10,22] & 121.24 (192) & [16,32] & 316.03 (288) & [6,26] & 184.89 (224) & [9,39] \\ 
High-perf & Galvatron (1F1B+Time)   & \textbf{226.73} (192) & [16,16] & 134.72 (96) & [24,24] & \textbf{349.72} (192) & [11,21] & 191.88 (128) & [17,31] \\ 
Cluster & Galvatron (1F1B+Bi-obj)  & \textbf{226.73} (192) & [16,16] & \textbf{144.59} (144) & [22,26] & \textbf{349.72} (192) & [11,21] & \textbf{207.08} (192) & [11,37] \\ 
\bottomrule
\end{tabular}%
}
\vspace{-3.5mm}
\end{table*}

\begin{table}[h]
\centering
\small
\caption{Comparison of Larger LLMs on 80GB GPUs.}
\vspace{-2mm}
\label{tab:llm_results}
\scalebox{0.85}{
{\begin{tabular}{c|ccc}
\toprule 
 \makecell{Strategy} & GPT3-15B & GPT3-39B & GPT3-65B \\
\midrule 
PyTorch DDP (DP)        & OOM  & OOM  & OOM\\ 
Megatron (TP)           & 12.31 (24)  & 5.56 (20)  & 2.02 (2) \\ 
PyTorch GPipe (PP)      & 26.56 (128)  & 12.38 (64)   & 4.62 (24) \\ 
FSDP/ZeRO-3 (SDP)       & 45.88 (128) & 14.31 (96)   & 3.40 (32) \\ 
DeepSpeed 3D            & 37.65 (128) & 13.79 (48) & 4.91 (16)\\ 
Galvatron (DP+TP)       & 14.36 (28)  & 6.82 (24)   & 2.99 (4)\\ 
Galvatron (DP+PP)       & 27.17 (72)  & 12.97 (48) & 4.62 (24)\\ 
Galvatron               & 50.79 (128) &   14.93 (96)  &   5.14 (16) \\ 
Galvatron (1F1B+Bi-obj) & 56.57 (1280) &  21.69 (768) &  10.62 (512)\\ 
Galvatron-Base          & 57.33 (512) &  21.14 (160) &  \textbf{13.97} (288)\\ 
Alpa                    & 45.51 (1024) &   13.75 (512)  &   7.16 (512) \\ 
Galvatron-BMW           & \textbf{61.42} (2560) & \textbf{22.86} (2048) & \textbf{13.97} (288)\\ 

\bottomrule
\end{tabular}}
}
\vspace{-6mm}
\end{table}

For example, \name enlarges the batch size from 520 to 960 for ViT-Huge-32 under 16 GPUs with 16 GB memory, and the throughput increases from 89.22 to 179.89 samples per second. The 2.02$\times$ speedup comes from the flexible fine-grained layer-level parallelism strategy, which helps to reduce the communication costs and improve the training efficiency. 
{As can be seen, \name outperforms all baselines across both two clusters as well as different models (including imbalanced model T5-512/4). This highlights \name's adaptive capability when handling different clusters and diverse model workloads.}
We then extend to an larger GPU cluster including 64 Nvidia A100 40GB GPUs, where each server has 8 GPUs equipped with NVLink and the servers are connected by 100 Gb InfiniBand network. Since the environment scale is significantly larger than before, we also increase the model sizes to 10 billion parameters (i.e., BERT-xHuge and ViT-xHuge, details are in Table~\ref{tab:model_config}). As we can see in Table~\ref{tab:64gpus_results}, even on such a large GPUs cluster, \name still outperforms these baseline methods.
Besides, based on our observations, the search time costs do not exponentially grow (i.e., 2.2$\times$ and 9.2$\times$ for 16 GPUs and 64 GPUs respectively compared with 8 GPUs), which is still tolerable. 


We further verify the scalability to model size by experimenting on larger LLMs, GPT-3 15B, 39B and 65B.
Given the substantial size of these LLMs, we compare on GPUs with larger memory capacity and higher communication bandwidth (32 Nvidia A100 80GB GPUs connected by 400Gb InfiniBand).
We also compare with Alpa~\cite{DBLP:conf/osdi/ZhengLZZCHWXZXG22} which automates parallelism on JAX.  
As shown in Table~\ref{tab:llm_results}, \name's performance enhancement is more pronounced with larger models.
Alpa, despite its broader operator-level search space, primarily treats SDP as a secondary optimization, allowing only DP or SDP, but not both, across the entire model, thus missing a more nuanced DP-SDP trade-off. Furthermore, 
CKPT proves a critical dimension for larger models under resource constraints, which is not considered in Alpa's search space.
In contrast, \name effectively balances memory, computation, and communication, delivering superior performance consistently.


\vspace{-3mm}
{
\subsection{Ablation Study}
\vspace{-1.0mm}
To delve deeper into the performance of bi-objective optimization of workload balance, we conduct experiments on different pipeline partitions. The results of BERT and T5-512/4 on high-performance cluster (16 A100) are shown in \autoref{tab:bi_obj_results}.
The memory-balanced partition can evidently support the training of larger batch sizes for the provided models, enabling a batch size of 128 for T5-512/4-32 under 8GB memory, in contrast to the limit of 48 in time-balanced partition. On the other hand, the time-balanced partition encourages a more uniform workload distribution across varying pipeline stages. As an example, a balanced partition of [16,16] is accomplished for the BERT-Huge-32, which has homogeneous layers.
Indeed, bi-objective optimization harnesses the advantages of both memory- and time-balanced partition, striking an optimal balance between memory and time cost to yield superior performance. In the training of BERT-Huge-48 and T5-512/4-48 under a memory constraint of 16G, the pipeline partitions are tuned to intermediary configurations of [22, 26] and [11, 37], while the training batch sizes are optimized as 144 and 192. These strategic adaptations culminate in marked improvements in system throughput, with surges of up to 19\% and 12\%,
thus underscoring the effectiveness of bi-objective optimization.
}

\vspace{-3mm}
\subsection{Optimal Parallelism Plan}
\vspace{-1mm}
{We list some examples of the optimal parallelism plans suggested by \name in \autoref{fig:opt_strategy}. For comparison, we choose different models (BERT-Huge-32, Swin-Huge-32, and T5-512/4-32), different GPU numbers (8 and 16), and different GPU clusters.} {
In case A for BERT, \name provides an optimal strategy containing $S_1^A$, PP+DP+CKPT, and $S_2^A$, PP+SDP+CKPT, incorporating SDP and CKPT to reduce memory costs and enlarge the batch size as well as the throughput. In case B for Swin-Huge-32, the optimal plans given by \name is rather complex, as it has four different layers which have different strategy preference. In Swin Transformer, shallower layers have larger activation size and smaller parameter size. 
To reduce memory consumption and communication overhead, shallower layers prefer data parallel which splits input activations and communicates parameter gradients, while deeper layers prefer tensor parallel which splits model parameters and communicates activations. Therefore, by mixing PP+DP, PP+DP+CKPT and PP+TP+DP, \name maximizes the memory utilization as well as the system efficiency.}
{
To further analyze \name, in case C, we test T5-512/4-32 on 16 low-performance GPUs and 16 high-performance GPUs under the same memory budget. In T5-512/4, as the decoder has shorter sequence length than the encoder, the activation size of the decoder is much less than the encoder, while the parameter size of the decoder is larger than the encoder due to the extra attention block. Therefore, similar to Swin, on low-performance GPUs, \name prefers TP for the decoder to reduce communication volume, where $S_5^C$, PP+DP+TP, is recommended. However, on high-performance GPUs, the advantage of low communication volume is not significant due to the high communication bandwidth, and \name mixes PP+SDP+CKPT, PP+SDP and PP+DP, preferring using SDP to reduce the memory consumption.
}

\vspace{-2mm}
\section{Conclusion}
In this work, we presented \name, a novel automatic parallel Transformer training system over multiple GPUs. Through the carefully designed search space decomposition and exploration algorithm, \name significantly outperforms the state-of-the-art baselines on the training throughput. We hope the open source release of \name will facilitate the future research directions on more challenging scenarios, e.g., heterogeneous environments and large DL models with complex and dynamic structures.

\vspace{-2mm}
\section*{Acknowledgments}
This work is supported by National Key R\&D Program of China (2022ZD0116315), NSFC (U23B2048, U22B2037), China National Postdoctoral Program for Innovative Talents (BX20230012), Beijing R\&D Program (Z231100010323002), Beijing NSF (4244080), and ZTE-PKU joint program. B. Cui, F. Fu and X. Miao are the co-corresponding authors.

\vspace{-2mm}
\normalem
\bibliographystyle{IEEEtran}      
\bibliography{my-reference}

\clearpage
\newpage
\newpage
\appendix

\subsection{Dynamic Programming}
\label{appendix:dp}
\subsubsection{State Transition of the Memory Consumption}
\label{appendix:memory_state_transition}
To optimize the total execution time $C(L, E)$ given the memory constraint $E_{all}(L)$ $\leq E$, the overall memory consumption $E_{all}(L)$ needs to be transitioned from $E_{all}(L-1)$ as follows: $E_{all}(L) = \max\{E_{all}(L-1), E_{f}(L-1)+ O_f(L, S_j)+O_b(L, S_j)+O_{ms}(L, S_j)\}$. We find that due to the maximum operation in Eq.~\ref{eq:E_overall_Sj}, the forward memory consumption $E_{f}(L-1)$ also need to be stored for state transition, therefore the states of dynamic programming need to be designed as $C(L, E_{all}, E_{f})$. Such states lead to a quadratic complexity in terms of memory constraint $E$, which is unacceptable in practice. Therefore, we first optimize $C(L, E_{fwd})$ with the forward memory constraint $E_{f}(L) \leq E_{fwd}$, where $E_{fwd} \leq E$, and finally check the validity of the overall memory, i.e., $E_{all}(L)\leq E$. The state transition of the forward memory consumption is quite simple: $E_{f}(L)=E_{f}(L-1)+O_f(L, S_j)+O_{ms}(L, S_j)$. This only requires one memory state to be stored for state transition, resulting in a linear complexity in terms of memory constraint.
\subsubsection{Proof for Optimal Substructure Property}
\label{appendix:optimal_substructure}
Before applying the dynamic programming in Section~\ref{sec:dynamic_programming}, we first prove that the problem follows the optimal substructure property. To obtain the minimum execution time $C(L, E_{fwd})$, we clarify that the solution must contain the sub-problem solution $C(L^{'}, E_{fwd}^{'})$, which represents the minimum execution time for the sub-model, i.e., first $L^{'}$ layers ($L^{'}\leq L$), within a smaller forward memory budget $E_{fwd}^{'}$ ($E_{fwd}^{'}\leq E_{fwd}$). This clarification holds because if the optimal solution $C(L, E_{fwd})$ does not contain a specific $C(L^{'}, E_{fwd}^{'})$, we can always reduce the total execution time by replacing the sub-problem solution to $C(L^{'}, E_{fwd}^{'})$. Due to the linear sequence model structure, the parallelization plan of the first $L^{'}$ layers will not affect the rest $L-L^{'}$ layers given the same memory budget $E_{fwd}-E_{fwd}^{'}$. Therefore, the problem satisfies the optimal substructure property for dynamic programming.
\subsubsection{Dynamic Programming Algorithm}
\label{appendix:dp_algorithm}
Based on the state transition formula in Eq.~\ref{eq:search_new}, we illustrate our overall dynamic programming algorithm in Algorithm~\ref{alg:DP}. Given a device memory budget $E$, we use $E_{fwd}$ ($E_{fwd} \leq E$) as the forward memory budget, and the rest part $E-E_{fwd}$ is spared for the backward peak memory $E_{b}(L)=E_{all}(L)-E_{f}(L)$. To maximize the memory utility, we gradually increase and traverse $E_{fwd}$ (line 11) to optimize $C(L, E_{fwd})$ using state transition formula in Eq.~\ref{eq:search_new} (line 14).
Then, we calculate the overall memory consumption $E_{all}(L)$ with the searched strategies $\mathcal{S}$ (line 18), and check the validity of the overall memory (line 19), i.e., $E_{all}(L)\leq E$. 
If the overall memory doesn't exceed $E$, we continue the optimization with a larger $E_{fwd}$.
Finally, we find the largest forward memory budget $E_{fwd}^{opt}$ with its searched strategies $\mathcal{S}^{opt}$ satisfying $E_{all}(L)\leq E$, and the optimized throughput $C(L, E_{fwd}^{opt})$ as well as the optimal strategy $\mathcal{S}^{opt}$ are the final output (line 26). 
Besides, to avoid unnecessary checks, we calculate an upper bound of the backward peak memory $E_{b}(L)$, $b_{up} = \max_{l=1}^{L} \max_{S_j\in S} O_{b}(l,S_j)$, and only verify its validity when $E-b_{up} < E_{fwd} \leq E$, as it can be easily proved that it satisfies $E_{all}(L) \leq E$ when $E_{fwd} \leq E-b_{up}$.

Note that, given the searched strategies, the calculation of $E_{all}(L)$ with Eq.~\ref{eq:E_overall_Sj} has $\mathcal{O}(L)$ complexity, while the state transition for all layers (line 13-15) requires $\mathcal{O}(L|S|)$ complexity. Therefore, the validity check does not contribute additional time complexity, and the complexity of dynamic programming algorithm~\ref{alg:DP} is $\mathcal{O}(LE|S|)$.
\begin{algorithm}[t]
\scalebox{0.92}{
\begin{minipage}{1.0\textwidth}
\caption{Dynamic Programming Algorithm}
\label{alg:DP}
\LinesNumbered
\KwIn{device memory: $E$, model stage: $M$,  \\
\quad \quad \quad batch size: $B$, strategy set: $S$}
\KwOut{minimum time cost $C$}
$L \gets length(M) $\;
\For{$l \gets 1, 2, ... L$}{
    \For{$S_j \in S $}{
        $O_{f}(l,S_j), O_{b}(l,S_j), O_{ms}(l,S_j), c(l,S_j) \\
        \hangindent=1.0em
        \hangafter=1
        \gets CostModel(M[l], B, S_j)$\;
    }
}
$C(0,\cdot) \gets 0, C(\cdot,0) \gets \infty$\;
Optimal strategy list $\mathcal{S} \gets []$\;
$b_{up} \gets \max_{l=1}^{L} \max_{S_j\in S} O_{b}(l,S_j)$\;
$E_{fwd}^{opt} \gets E-b_{up}$\;
\For{$E_{fwd} \gets 1, 2, ... E$}{
    Strategy list $\mathcal{S} \gets []$\;
    \For{$l \gets 1, 2, ... L$}{
        $C(l,E_{fwd}), S_{min} \gets $ Transition formula in Eq.~\ref{eq:search_new}\;
        $\mathcal{S}$\texttt{.append}($S_{min}$)\; 
    }
    \If{$E_{fwd}>E-b_{up}$}{
        Given $\mathcal{S}$, calculate $E_{all}(L)$ with Eq.~\ref{eq:E_overall_Sj}\;
        \eIf{$E_{all}(L) \leq E$}{
            $E_{fwd}^{opt} \gets E_{fwd}$, $\mathcal{S}^{opt} \gets \mathcal{S}$\;
        }{
            \textbf{break}\;
        }
    }
}
\textbf{return} $C(L,E_{fwd}^{opt}), \mathcal{S}^{opt}$\;
\end{minipage}
}
\end{algorithm}

\subsection{Pipeline Partition Adjustment}
\label{appendix:pp_adjustment}
We aim to adjust the pipeline partition $\mathbf{p}$ from memory-balanced to time-balanced. Given the memory-balanced partition $\mathbf{p}_m$ and time-balanced partition $\mathbf{p}_t$, we initialize $\mathbf{p}$ as $\mathbf{p}_m$. In each iteration, when adjusting $\mathbf{p}$ to obtain a new partition $\mathbf{p}'$, the three limitations listed in Section~\ref{subsection:bi_obj_workflow} should be satisfied. Based on the definition of $\alpha_t$ and $\alpha_m$ in Eq.~\ref{eq:balance_deg}, limitation~1 ensures $\alpha_t(\mathbf{p}') \geq \alpha_t(\mathbf{p})$, as the maximum stage time cost of $\mathbf{p}'$ is lower than $\mathbf{p}$. Furthermore, limitation~2 and 3 ensure $\alpha_m(\mathbf{p}') \geq \alpha_m(\mathbf{p}_t)$, as the maximum stage memory cost of $\mathbf{p}'$ is lower than $\mathbf{p}_t$. Therefore, we can demonstrate that:
\begin{equation}
\label{eq:adjust}
\begin{aligned}
 \alpha_t(\mathbf{p}_m) \leq \alpha_t&(\mathbf{p}) \leq \alpha_t(\mathbf{p}') \leq \alpha_t(\mathbf{p}_t), \\
 \alpha_m(\mathbf{p}_t) \leq &\alpha_m(\mathbf{p}') \leq \alpha_m(\mathbf{p}_m).
\end{aligned}
\end{equation} 
\noindent which indicates that $\mathbf{p}'$ is a potentially optimal partition as it satisfies condition~\ref{eq:inter_state_def}, and $\mathbf{p}'$ is superior to the prior partition $\mathbf{p}$ in terms of time balance, which is the target of our adjustment. Then, $\mathbf{p}'$ is pushed to the queue for subsequent search iterations.

\subsection{Pipeline Cost Estimator}
\label{appendix:pp_cost_estimator}
In this subsection, we introduce our pipeline cost estimator in detail.
By summing up the layer cost $c(l,s)$ for each layer (Section~\ref{subsection:cost_model}), we calculate the cost of pipeline model stage $M_i$, $C(M_i,B_m)$, where $B_m$ is the micro-batch size, and $m$ is the micro-batch number. We then estimate the overall time cost of the model according to Eq.~\ref{eq:pipeline_cost_imbalance} as follows:
\vspace{-1mm}
\begin{equation}
\label{eq:overall_costmodel}
\scalebox{0.92}{$
\begin{aligned}
C(M,B) &= (m - 1) * \max \limits_{i=1}^{P}C_{no\_grad\_sync}(M_i,B_m) \\[-2.0ex]
&\quad + \sum\limits_{i=1}^{P}C(M_i,B_m)
\end{aligned}
$}
\vspace{-2mm}
\end{equation}

\noindent where $C(M_i,B_m)$ denotes the execution time of model stage $M_i$ under micro-batch size $B_m$ ($B_m = B / m$), and $C_{no\_grad\_sync}(M_i,B_m)$ denotes the execution time of stage $M_i$ without gradient synchronization.

The workload balance is considered in Eq.~\ref{eq:overall_costmodel},
and we also consider the different execution time between the last micro-batch and others. Due to gradient accumulation, data parallelism requires gradient synchronization during back propagation of the last micro-batch, causing longer execution time for the last micro-batch than previous ones. Therefore, we use $C(M_i,B_m)$ to simulate the cost of the last micro-batch, which contains the gradient communication cost, and use $C_{no\_grad\_sync}(M_i,B_m)$ to simulate that of the previous ones, where we remove the communication cost of gradient synchronization. 
In conclusion, our cost model is both efficient and accurate, which allows us to optimize the pipeline efficiency and promote the system performance.


\begin{figure}[h]
\centering
\begin{minipage}[t]{0.48\linewidth}
\includegraphics[width=1.0\linewidth]{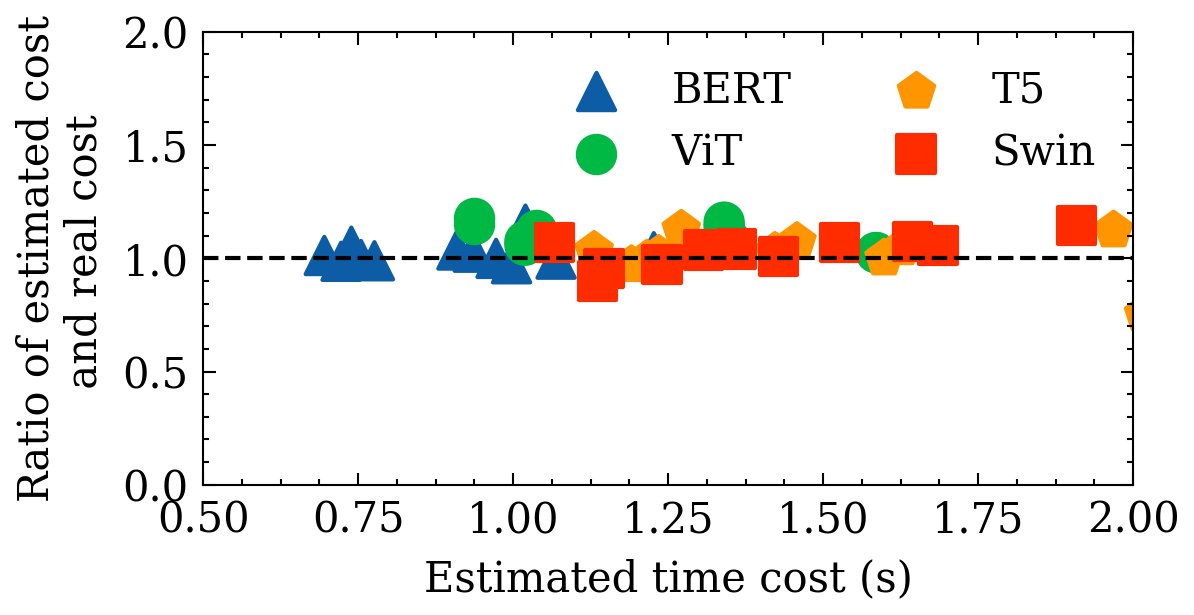}

\subcaption{w. overlapping slowdown}
\end{minipage}
\begin{minipage}[t]{0.48\linewidth}
\includegraphics[width=1.0\linewidth]{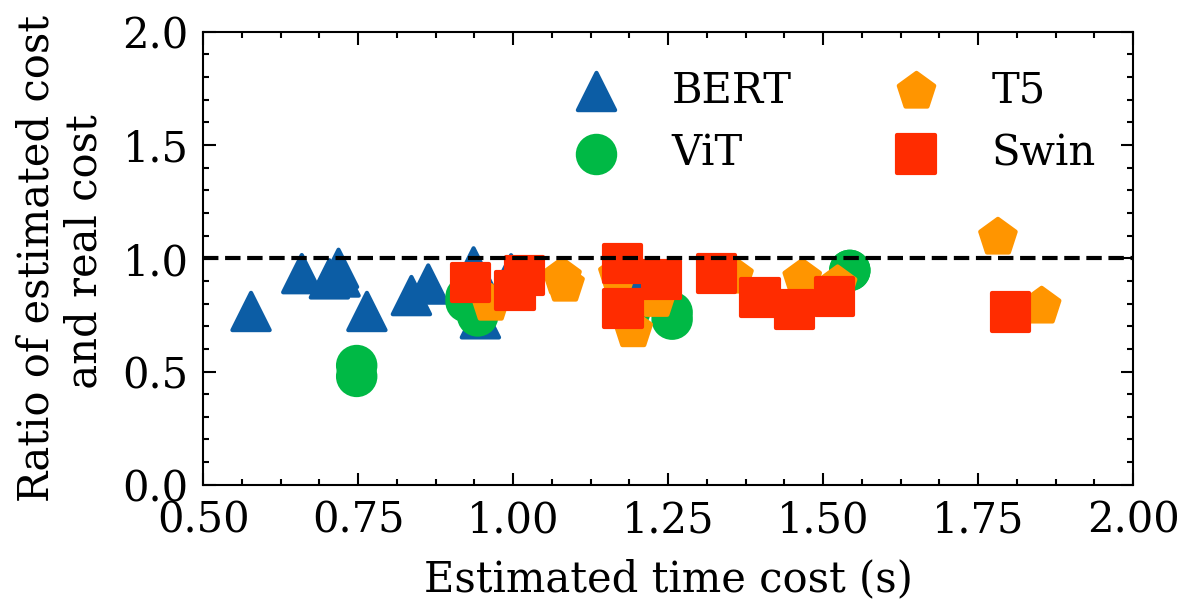}

\subcaption{w.o. overlapping slowdown}
\end{minipage}

\caption{Estimation errors with and without considering the overlapping slowdown.}\label{fig:costestimation}
\vspace{-2mm}
\end{figure}

\subsection{Estimation Performance}
In this subsection, we measure the performance of our cost estimator.
\autoref{fig:costestimation} demonstrates the cost estimation errors with and without considering the overlapping slowdown. It can be observed that our estimation results are very close to the real execution costs for all experimental models. The average prediction error is less than 5\%. However, when ignoring the slowdown, the estimations become obviously lower, resulting in an average prediction error of more than 15\%, which compromises the efficiency of the generated execution strategy.

\end{document}